\documentclass[11pt,a4paper]{article}
\usepackage[hyperref]{acl2020}
\usepackage{times}
\usepackage{latexsym}

\usepackage{multicol}
\usepackage{subcaption}

\usepackage{microtype}

\aclfinalcopy % Uncomment this line for the final submission
\usepackage[shortlabels]{enumitem}
\usepackage{mathtools}
\usepackage{amssymb}
\usepackage{url}
\usepackage{booktabs}
\usepackage{xcolor}
\usepackage{mdframed}
\usepackage{framed}
\usepackage{comment}
\usepackage{listings}
\usepackage{float}
\definecolor{LightBlue}{RGB}{239,247,255}
\definecolor{DarkBlue}{RGB}{19,10,200}
\definecolor{DarkGreen}{RGB}{60,125,0}

\newtheorem{mdtheoremm}{Example}
\newenvironment{example}%
  {\begin{mdframed}[backgroundcolor=LightBlue]\begin{mdtheoremm}}%
  {\end{mdtheoremm}\end{mdframed}}

\newmdenv[innerlinewidth=20pt, roundcorner=20pt,linecolor=DarkBlue,innerleftmargin=3pt,
innerrightmargin=3pt,innertopmargin=3pt,innerbottommargin=3pt]{mybox}

\usepackage{tcolorbox}

\definecolor{DarkRed}{RGB}{130,25,0}
\definecolor{DarkGreen}{RGB}{30,130,30}

\usepackage[colorinlistoftodos]{todonotes}

\newcommand{\ignore}[1]{}

\newcommand{\alg}{S}
\newcommand{\measure}[1]{\mathcal{M}(#1)}

\newcommand{\hn}{$H_0$}
\newcommand{\ha}{$H_1$}
\newcommand{\prob}[1]{\mathbb{P}(#1)}
\newcommand{\highlight}[1]{{\color{red}#1}}

\usepackage{array}
\newcommand{\PreserveBackslash}[1]{\let\temp=\\#1\let\\=\temp}
\newcolumntype{R}[1]{>{\PreserveBackslash\raggedleft}p{#1}}
\newcolumntype{L}[1]{>{\PreserveBackslash\raggedright}p{#1}}

\newcolumntype{C}[1]{>{\centering\arraybackslash} m{#1}}

\newcommand\T{\rule{0pt}{2.6ex}}       % Top strut
\usepackage{xspace}

\newcommand{\hybayes}{\textbf{\texttt{HyBayes}}\xspace}

\title{
\vspace*{-0.5in}
{{\small \hfill ACL'20}\\
\vspace*{.25in}} 
\emph{Not All Claims are Created Equal:} \\ 
Choosing the Right Statistical Approach to Assess Hypotheses
}
\author{
Erfan Sadeqi Azer$^{1}$ \; Daniel Khashabi$^{2}$\thanks{\ \ Work done while the second author was affiliated with the University of Pennsylvania.} \; Ashish Sabharwal$^{2}$ \; Dan Roth$^{3}$  \\
 $^1$Indiana University \;  $^2$Allen Institute for Artificial Intelligence  \; $^3$University of Pennsylvania  \\
 {\tt  \footnotesize
 esadeqia@indiana.edu \; 
 \{danielk,ashishs\}@allenai.org \;
 danroth@cis.upenn.edu
 }  
}

\date{}

\begin{document}
\maketitle

\begin{abstract}
Empirical research in Natural Language Processing (NLP) has adopted a narrow set of principles for assessing hypotheses, relying mainly on $p$-value computation, which suffers from several known issues. While alternative proposals have been well-debated and adopted in other fields, they remain rarely discussed or used within the NLP community. We address this gap by contrasting various hypothesis assessment techniques, especially those not commonly used in the field (such as evaluations based on Bayesian inference). Since these statistical techniques differ in the hypotheses they can support, we argue that practitioners should first decide their target hypothesis before choosing an assessment method. This is crucial because common fallacies, misconceptions, and misinterpretation surrounding hypothesis assessment methods often stem from a discrepancy between what one would like to claim versus what the method used actually assesses. Our survey reveals that these issues are omnipresent in the NLP research community. As a step forward, we provide best practices and guidelines tailored towards NLP research, as well as an easy-to-use package called \hybayes for Bayesian assessment of hypotheses,\footnote{\label{footnote:toolkit}\url{https://github.com/allenai/HyBayes}} complementing existing tools.
\end{abstract}

%%%%%%%%%%%%%%%%%%%%%%%%%%%%%
\section{Introduction}

% \newcommand\blfootnote[1]{%
%   \begingroup
%   \renewcommand\thefootnote{}\footnote{#1}%
%   \addtocounter{footnote}{-1}%
%   \endgroup
% }

% \blfootnote{* Work was done while the second author was affiliated with the University of Pennsylvania.}

Empirical fields, such as Natural Language Processing (NLP), must follow scientific principles for assessing hypotheses and drawing conclusions from experiments. For instance,
suppose we come across the results in Table~\ref{tab:table_output_dataset}, summarizing the accuracy of two question-answering (QA) systems $S_1$ and $S_2$ on some datasets. What is the correct way to interpret this empirical observation in terms of the superiority of one system over another? While $S_1$ has higher accuracy than $S_2$ in both cases, the gap is moderate and the datasets are of limited size. 
% Could this apparent accuracy difference be just result a random chance? 
% Could this apparent difference in accuracy only be just a random chance? 
Can this apparent difference in performance be explained simply by random chance, or do we have sufficient evidence to conclude that $S_1$ is in fact \emph{inherently different} (in particular, inherently stronger) than $S_2$ on these datasets? If the latter, can we quantify this gap in inherent strength while accounting for random fluctuation?

\begin{table}[t]
    \centering
    \setlength{\tabcolsep}{2pt}
    \footnotesize
    % \scalebox{0.76}{
    \begin{tabular}{cc|cc|cc}
        \toprule
        System & Description & \multicolumn{2}{c}{ ARC-easy } & \multicolumn{2}{c}{ ARC-challenge } \\
        ID & & \#Correct & Acc. & \#Correct & Acc. \\
        \midrule
        $S_1$ & BERT & 1721 & 72.4 & 566 & 48.3 \\
        $S_2$ & Reading Strategies & 1637 & 68.9 & 496 & 42.3 \\
        \bottomrule
    \end{tabular}
    % }
    % \vspace{-0.2cm}
    \caption{Performance of two systems~\cite{devlin2019bert,sun2018improving} on the ARC question-answering dataset~\cite{Clark2018ThinkYH}. ARC-easy \& ARC-challenge have $2376$ \& $1172$ instances, respectively. Acc.: accuracy as a percentage.}
    \label{tab:table_output_dataset}
\end{table}

Such fundamental questions arise in one form or another in every empirical NLP effort. Researchers often wish to draw conclusions such as:

    \begin{tcolorbox}[
                colback=LightBlue,colframe=blue!125!black,
                  left=0.5pt,
                  right=0.5pt,
                  top=0.5pt, 
                  bottom=0.5pt, 
                  boxrule=0.1mm
                  ]
        \emph{
        \small
        \begin{enumerate}[(Ca), topsep=0.1pt,itemsep=0.1ex,partopsep=0.1ex,parsep=0.1ex,leftmargin=*, font=\bfseries]
            \footnotesize
            \item I'm 95\% \highlight{confident} that $S_1$ and $S_2$ are inherently \highlight{different}, in the sense that if they were inherently identical, it would be highly unlikely to witness the observed 3.5\% empirical gap for ARC-easy.
            \item With \highlight{probability} at least 95\%, the inherent accuracy of $S_1$ \highlight{exceeds} that of $S_2$ by at least 1\% for ARC-easy.
        \end{enumerate}
        }
    \end{tcolorbox}
    
These two conclusions differ in two respects. First, \emph{\textbf{Ca}} claims the two systems are inherently different, while \emph{\textbf{Cb}} goes further to claim a margin of at least 1\% between their inherent accuracies. The second, more subtle difference lies in the interpretation of the 95\% figure: the 95\% confidence expressed in \emph{\textbf{Ca}} is in terms of the space of empirical observations we could have made, given some underlying truth about how the inherent accuracies of $S_1$ and $S_2$ relate; while the 95\% probability expressed in \emph{\textbf{Cb}} is directly over the space of possible inherent accuracies of the two systems. 

% Incidentally, turning these desired claims into proper mathematical statements isn't completely straightforward. We can, for example, turn these into
To support such a claim, one must turn it into a proper mathematical statement that can be validated using a statistical calculation.
This in turn brings in additional choices: we can make at least
four statistically \emph{distinct} hypotheses here, each supported by a different statistical evaluation: 

% \begin{table}[]
% \centering
    % \scalebox{0.99}{
    \begin{tcolorbox}[
                colback=LightBlue,colframe=blue!125!black,
                  left=0.5pt,
                  right=0.5pt,
                  top=0.5pt, 
                  bottom=0.5pt, 
                  boxrule=0.1mm
                  ]
        \emph{
        \small
        \begin{enumerate}[(H1), topsep=0.1pt,itemsep=0.1ex,partopsep=0.1ex,parsep=0.1ex,leftmargin=*, font=\bfseries]
            \footnotesize
            \item \highlight{Assuming} $S_1$ and $S_2$ have inherently \highlight{identical} accuracy, the probability (\textbf{p-value}) of making a hypothetical observation with an accuracy gap at least as large as the empirical observation (here, 3.5\%) is at most 5\% (making us 95\% confident that the above assumption is false).
            % it would contradict the assumption of $S_1$ and $S_2$ having identical inherent accuracy.) 
            %
            % \item 
            % The 95\% confidence interval consists of all values of the inherent accuracies that would not be rejected by a (two-tailed) significance test that allows 5\% false alarms.
            %
            % How much more extreme a hypothetical observation could be (compared to the given empirical observations), such that the probability of such hypothetical observations is not too small.
            % how much can their performance gap deviate from the observed values
            % 
            % ASHISH'S VERSION: 
            % \item \highlight{Assuming} $S_1$ and $S_2$ have inherently \highlight{identical} accuracy,
            % let $G$ be the largest value (\textbf{confidence interval}) such that the probability of observing a gap of at least $G$ is over 5\% (making gaps up to $G$ attributable to random chance). Then $G <$ 3.5\%, implying the observed gap is more than just random fluctuation.
            % 
            \item            \highlight{Assuming} $S_1$ and $S_2$ have inherently \highlight{identical} accuracy, the empirical accuracy gap (here, 3.5\%) is larger than the maximum possible gap (\textbf{confidence interval}) that could hypothetically be observed with a probability of over 5\% (making us 95\% confident that the above assumption is false).
            %\highlight{Assuming} $S_1$ and $S_2$ have inherently \highlight{identical} accuracy, 
            % let $G$ denote the largest possible accuracy gap (\textbf{confidence interval}) that could hypothetically be observed with probability over 5\% (making gaps up to $G$ attributable to random chance). If the empirical gap (here, 3.5\%) lies within $G$ it implies that the observed gap is more than just random fluctuation.
            % If we had started with the assumption that $S_1$ is better than $S_2$ by exactly \highlight{$x$ margin}, for which values of $x$, the probability in {\textbf{H1}} would be bigger than a predefined threshold. 
            % test in (a) would not be rejected.
            % \item Prior to any empirical observations, suppose we hold a belief with respect to the relative strength of the two systems. After these empirical observations, how probable is that the inherent accuracy of $S_1$ is strictly higher than $S_2$ by a margin of $x\%$?
            % \item Given a Prior belief with respect to the relative strength of the two systems as well as these empirical observations, how probable is that the inherent accuracy of $S_1$ is strictly higher than $S_2$ by a margin of $x\%$?
            %
            \item \highlight{Assume a prior belief} (a probability distribution) w.r.t.\ the inherent accuracy of typical systems. Given the empirically observed accuracies, \highlight{the probability} (\textbf{posterior interval}) that the inherent accuracy of $S_1$ exceeds that of $S_2$ by \highlight{a margin} of 1\% is at least 95\%.
            \item \highlight{Assume a prior belief} (a probability distribution) w.r.t.\ the inherent accuracies of  typical systems. Given the empirically observed accuracies, \highlight{the odds} increase by a factor of 1.32 (\textbf{Bayes factor}) in favor of the hypothesis that the inherent accuracy of $S_1$ exceeds that of $S_2$ by \highlight{a margin} of 1\%.
            % \item \highlight{Assume a prior belief} (a probability distribution) w.r.t.\ the relative inherent accuracies of $S_1$ and $S_2$. Given the empirically observed accuracies, \highlight{the odds} increase by a factor of 1.32 (\textbf{Bayes factor}) in favor of the hypothesis that the inherent accuracy of $S_1$ exceeds that of $S_2$ by \highlight{a margin} of 1\%.
            % \item \highlight{Assume a prior belief} (a probability distribution) w.r.t.\ the relative inherent accuracies of $S_1$ and $S_2$. Given empirical observations, \highlight{the odds} (\textbf{Bayes factor}) increase by 95\%(?) in favor of the hypothesis that the inherent accuracy of $S_1$ exceeds that of $S_2$ by \highlight{a margin} of 1\%.
            %
        \end{enumerate}
        }
    \end{tcolorbox}
    % }
%     \vspace{-0.7cm}
%     \caption{.}
%     \label{tab:example:hypotheses}
% \end{table}
% \todo{clarify the phrasing of H2}

% We use the task of answering natural language questions~\cite{Clark2018ThinkYH} as our running example and compare multiple systems.\footnote{More details at: https://leaderboard.allenai.org/arc/} For this task, 
% the metric $\measure{\alg, \boldsymbol{x}}$ is defined as a binary function which denotes whether a system answers a given question correctly or not  
% Each system is credited with the observed ratio of questions answered correctly, that forms $\boldsymbol{y}$ 

% \todo{
%     DanielK: post-ACL submission: H3 used to have a margin parameter $x$ which somehow got dropped. We refer to this parameter in Example 3, but unfortunately this parameter doesn't exist anymore. 
% }

As this illustrates, there are multiple ways to formulate empirical hypotheses and support empirical claims. Since each hypothesis starts with a different assumption and makes a (mathematically) different claim, it can only be tested with a certain set of statistical methods.
% And each hypothesis could be tested with only a certain set of algorithms, since they
% start with different assumptions and provide results with different nature.
Therefore, \emph{NLP practitioners ought to define their target hypothesis before 
    % any effort to test or assess it.
    choosing an assessment method.
}
% Only after defining their hypothesis, a user can go for about choosing the corresponding statistical test. 

The most common statistical methodology used in NLP is null-hypothesis significance testing (NHST) which uses $p$-values~\cite{sogaard2014s,koehn2004statistical,dror2018recommended}. Hypotheses {\em\textbf{H1}\&\textbf{H2}} can be tested with $p$-value-based methods, which include confidence intervals and operate over the \emph{probability space of observations}\footnote{More precisely, over the probability space of an aggregation function over observations, called test statistics.}
(\S\ref{subsec:nhst:pvalue} and \S\ref{subsec:confidence:interval}).
On the other hand, there are often overlooked approaches, based on Bayesian inference~\cite{kruschke2018bayesian}, that can be used to assess hypotheses {\em\textbf{H3}\&\textbf{H4}} (\S\ref{subsec:posterior:interval} and \S\ref{subsec:bayes:factor}) and have two broad strengths: they can deal more naturally with accuracy margins and they operate directly over the \emph{probability space of inherent accuracy} (rather than of observations).

For each technique reviewed in this work, we discuss how it compares with alternatives and summarize common misinterpretations surrounding it (\S\ref{sec:comparisons}). 
For example, a common misconception about $p$-value is that it represents \emph{a probability of the validity of a hypothesis}.
While desirable, $p$-values in fact do not provide such a probabilistic interpretation (\S\ref{subsec:comparison:measures:of:uncertainty}). It is instead through a Bayesian analysis of the posterior distribution of the test statistic (inherent accuracy in the earlier example) that one can make claims about the probability space of that statistic, such as {\em\textbf{H3}}. 
% report sentences like the following ``Our experiments show that with $\alpha$ probability, the performance advantage of system-1 is at least $\beta$ units over system-2.'' 

We quantify and demonstrate related common malpractices in the field through a manual annotation of 439 ACL-2018 conference papers,\footnote{\url{https://www.aclweb.org/anthology/events/acl-2018/}} and a survey filled out by 55 NLP researchers (\S\ref{sec:common_practices}).
% We provide statistics on the common malpractices in the field. 
We highlight surprising findings from the survey, such as the following:
While 86\% expressed fair-to-complete confidence in the interpretation of $p$-values, only a small percentage of them correctly answered a basic $p$-value interpretation question.

% In order to choose an appropriate test, a user has to familiar with all of them and be able to distinguish what hypothesis they assess. However, our analysis of ACL'18 papers and a survey among NLP researcher, in \S\ref{sec:common_practices}, we observe that the vast majority of our community is not aware of alternative tests.  

% With increasing over-reliance on certain hypothesis testing techniques, there are growing troubling trends in misuse or misinterpretation of such techniques~\cite{goodman2008dirty,demvsar2008appropriateness}. We provide statistics on the common malpractices in the field. For instance, while 86\% expressed fair to complete confidence in interpreting $p$-values, yet only a small percentage of these could answer a $p$-value interpretation question correctly. 

% These practices encourage binary thinking: many papers (including many in the NLP community) over-rely on such binary decisions based on $p$-values to evaluate the merits of works. 
% In addition, other compounding factors like incomplete reporting of the tests~\cite{dror2018hitchhiker} have gradually diminished the effectiveness of scientific hypothesis tests. 
 
% Among the small portion of works that use any hypothesis assessment techniques, almost all are frequentist tests (such as $p$-values.) 
% However, our exposition contains techniques that are less utilized in the field. 

\paragraph{Contributions.}
% The main goal of this work is to 
This work seeks to inform the NLP community about crucial distinctions between various statistical hypotheses and their corresponding assessment methods, helping move the community towards well-substantiated empirical claims and conclusions. 
% \erfan{not testing. "Bayesian hypothesis testing" is only referred to Bayes factor analysis}
Our exposition covers a broader range of methods (\S\ref{sec:assessment}) than those included in recent related efforts (\S\ref{sec:related-work}), and highlights that these methods achieve different goals.
Our surveys of NLP researchers reveals problematic trends (\S\ref{sec:common_practices}), emphasizing the need for increased scrutiny and clarity. We conclude by suggesting guidelines for better testing (\S\ref{sec:recommended_practices}), as well as providing a toolkit called 
\hybayes
(cf.~Footnote~\ref{footnote:toolkit}) tailored towards commonly used NLP metrics.
% In summary, this work is expected to
We hope this work will
encourage an improved understanding of statistical assessment methods and effective reporting practices with measures of uncertainty.

\subsection{Related Work}
\label{sec:related-work}

While there is an abundant discussion of significance testing in other fields, only a handful of NLP efforts address it. For instance, \citet{chinchor1992statistical} defined the principles of using hypothesis testing in the context of NLP problems. 
Most-notably, there are works studying various randomized tests~\cite{koehn2004statistical,ojala2010permutation,graham2014randomized},  or metric-specific tests~\cite{evert2004significance}.
% \cite{yeh2000more} Proposes alternatives to consider the lack of independence between the instances. 
% \cite{koehn2004statistical} explains a bootstrap resampling test  in the context of machine translation. 
% \cite{evert2004significance}  a statistical model that interprets the evaluation of ranking methods 
% \cite{ojala2010permutation} permutation-based p-value 
% \cite{graham2014randomized} Randomized Significance Tests in Machine Translation 
More recently, \citet{dror2018hitchhiker} and \citet{dror2018recommended} provide a thorough review of \emph{frequentist} tests. 
While an important step in better informing the community, it covers
% a narrow layer of
a subset of statistical
tools. Our work complements this effort by pointing out alternative tests. 

With increasing over-reliance on certain hypothesis testing techniques, there are growing troubling trends of misuse or misinterpretation of such techniques~\cite{goodman2008dirty,demvsar2008appropriateness}. 
Some communities, such as statistics and psychology, even have published guidelines and restrictions on the use of $p$-values~\cite{editoral2015tarfimow,wasserstein2016asa}. 
In parallel, some authors have advocated for using alternate paradigms such as Bayesian evaluations~\cite{kruschke2010bayesian}.

NLP is arguably an equally empirical field, yet with a rare discussion of proper practices of scientific testing, common pitfalls, and various alternatives. In particular, while limitations of $p$-values are heavily discussed in statistics and psychology, only a few NLP efforts approach them: 
% pointed out any issues:
% potential limitations with $p$-value hypothesis testing: 
% the difference between estimates gained from randomized tests and 
% \cite{riezler2005some} that p-values are estimated more conservatively by approximate randomization than by bootstrap tests, thus increasing the likelihood of type-I error for the latter. We point out a pitfall of randomly assessing significance in multiple pairwise comparisons, and conclude with a recommendation to combine NIST with approximate randomization 
over-estimation of significance by model-based tests~\cite{riezler2005some}, 
lack of independence assumption in practice~\cite{berg2012empirical}, and sensitivity to the choice of the significance level~\cite{sogaard2014s}. 
% , are among a few work. 
% \cite{benavoli2016should}, 
% among others.
% \todo{methods~\cite{riezler2005some}, but we are not aware of any work on potential alternatives.  }
Our goal is to provide a unifying view of the pitfalls and best practices, and equip NLP researchers with Bayesian hypothesis assessment approaches as an important alternative tool in their toolkit.

% \paragraph{Bayesian model comparison.}
% \cite{benavoli2017time}
% \cite{benavoli2015bayesian}
% \cite{benavoli2014bayesian}
% \cite{demvsar2006statistical}
% \cite{corani2017statistical}

% \cite{bengio2004no}
% \cite{evert2001methods}
% \cite{chaudhari2011lexical}
% \cite{damani2013appropriately}
% \cite{sogaard2013estimating}
% \cite{dror2017replicability}

\section{Assessment of Hypotheses}
\label{sec:assessment}

We often wish to draw qualitative inferences based on the outcome of experiments 
% (for example, comparing the relative performance of systems.) 
% (for example, comparing characteristics of two systems.) 
(for example, inferring the relative inherent performance of systems). 
To do so, we usually formulate a hypothesis that can be \emph{assessed} through some analysis.

% A simple example of a
% test statistic is recall, which is a function of the number correct, the number partially correct, and the number of possible fills. 
Suppose we want to compare two systems on a dataset of instances $\boldsymbol{x} = [x_1, \hdots, x_n]$ with respect to a measure $\measure{\alg, x}$ representing the performance of a system $\alg$ on an instance $x$. 
Let $\measure{\alg, \boldsymbol{x}}$ denote the vector $[\measure{\alg,{x}_i}]_{i=1}^n$. 
Given systems $S_1, S_2$, define $\boldsymbol{y} \triangleq [\measure{S_1, \boldsymbol{x}} , \measure{S_2, \boldsymbol{x}} ]$ as a vector of observations.\footnote{For simplicity of exposition, we assume the performances of two systems are on a single dataset. However, the discussion also applies to observations on multiple different datasets.} 

\begin{figure}[t]
    \centering
    \includegraphics[trim=0.6cm 0.5cm 0cm 0.2cm, scale=0.27]{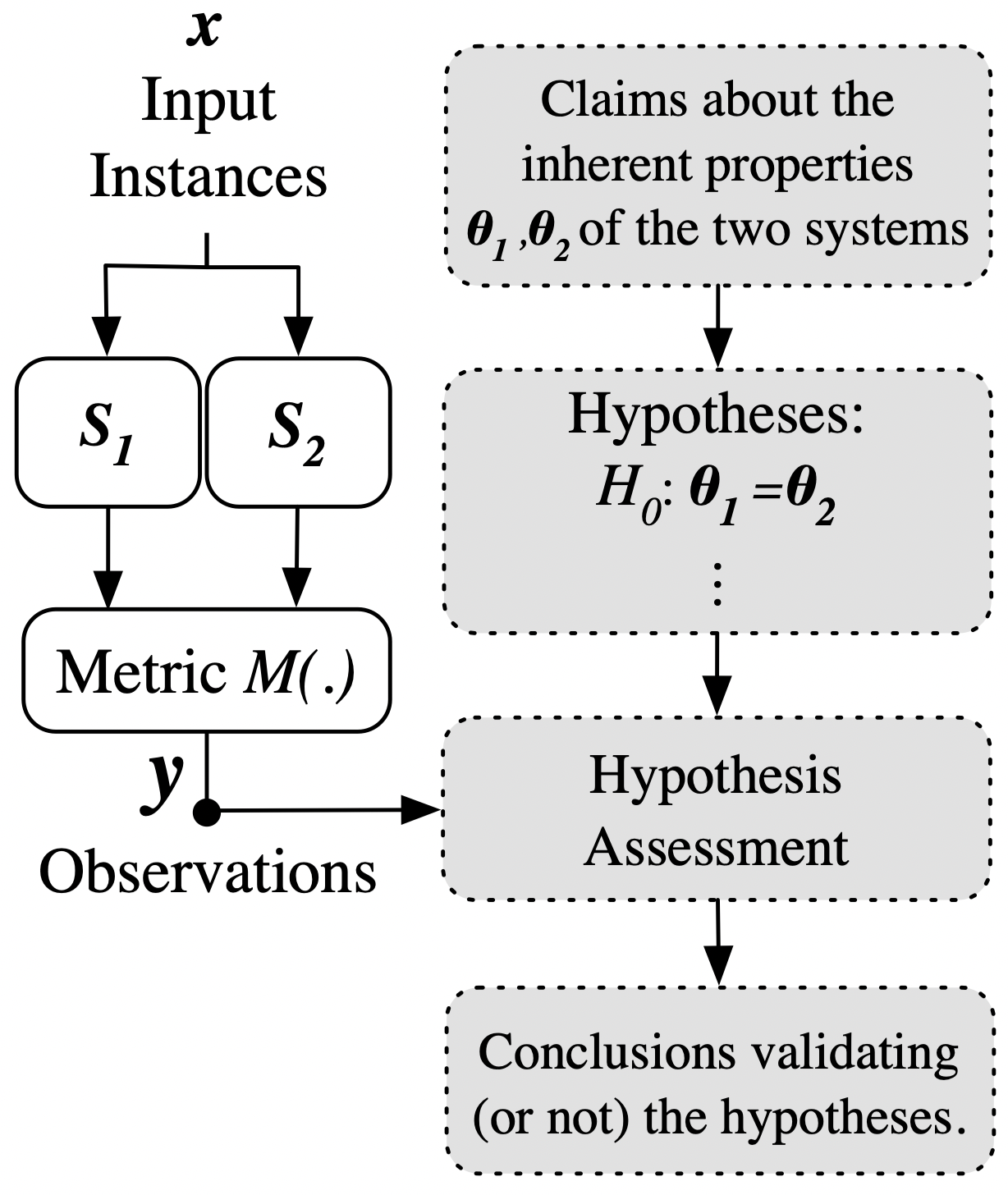}
    \caption{
        Progression of steps taken during a scientific assessment of claims from empirical observations. 
    }
    \label{fig:notation}
\end{figure}

In a typical NLP experiment, the goal is to infer some \emph{inherent} and \emph{unknown} properties of systems. To this end, a practitioner assumes a probability distribution on the observations $\boldsymbol{y}$, parameterized by $\boldsymbol{\theta}$, the properties of the systems. In other words, $\boldsymbol{y}$ is assumed to have a distribution\footnote{\emph{Parametric} tests assume this distribution, while \emph{non-parametric} tests do not.} with unknown parameters $\boldsymbol{\theta}$. In this setting, a \emph{hypothesis} $H$ is a condition on $\boldsymbol{\theta}$. Hypothesis assessment is a way of evaluating the degree to which the observations $\boldsymbol{y}$ are compatible with $H$. The overall
% configuration of the variables
process is depicted in Figure~\ref{fig:notation}. 

Following our running example, we use the task of answering natural language questions~\cite{Clark2018ThinkYH}. 
% and compare multiple systems.\footnote{More details at: https://leaderboard.allenai.org/arc/} 
While our examples are shown for this particular task, all the ideas are applicable to more general experimental settings.

For this task, the performance metric $\measure{\alg, x}$ is defined as a binary function indicating whether a system $S$ answers a given question $x$ correctly or not. The performance vector $\measure{\alg, \boldsymbol{x}}$ captures the system's accuracy on the entire dataset (cf.~Table~\ref{tab:table_output_dataset}). 
We assume that each system $S_i$ has an unknown \emph{inherent accuracy} value, denoted $\theta_i$. Let $\boldsymbol{\theta} = [\theta_1, \theta_2]$ denote the unknown inherent accuracy of two systems.
% }
% {\color{purple}
% On the other hand, each system has a latent and inherent accuracy denoted with $\theta_i$ (for the $i$th system). 
% }
% {
% \color{blue}
% In this setting $\boldsymbol{\theta} = [\theta_1, \theta_2]$, where $\theta_i$ denote such accuracy for $i$th system.
% }
% {
% \color{red}
% In this problem $\theta_i$ denote such accuracy for $i$th system. Then, the vector $\boldsymbol{\theta}$ introduced above has just two components in our example, i.e., $\boldsymbol{\theta}$. 
% }
In this setup, one might, for instance, be interested in assessing the credibility of the hypothesis $H$ that $\theta_1 < \theta_2$.
% While we perform our analysis on this particular example ???
% While we use this particular example 
% \todo{
% Jordan: 
% if this is true, is it a good idea to present this particular task as an example?  (also `task' is a very NLP thing to say!) 
% }
% none of these ideas are limited to this problem definition. 
% While we demonstrate techniques on this problems, it should be noted that none of these ideas are limited to this problem definition. 
% \todo{Put the table of observation from the spread sheet: rows: algorithms, 4 columns: 2 for each dataset.}

\begin{table}[ht]
    \centering
    \includegraphics[scale=0.18,trim=0.5cm 1.0cm 0cm 0.6cm, clip=false]{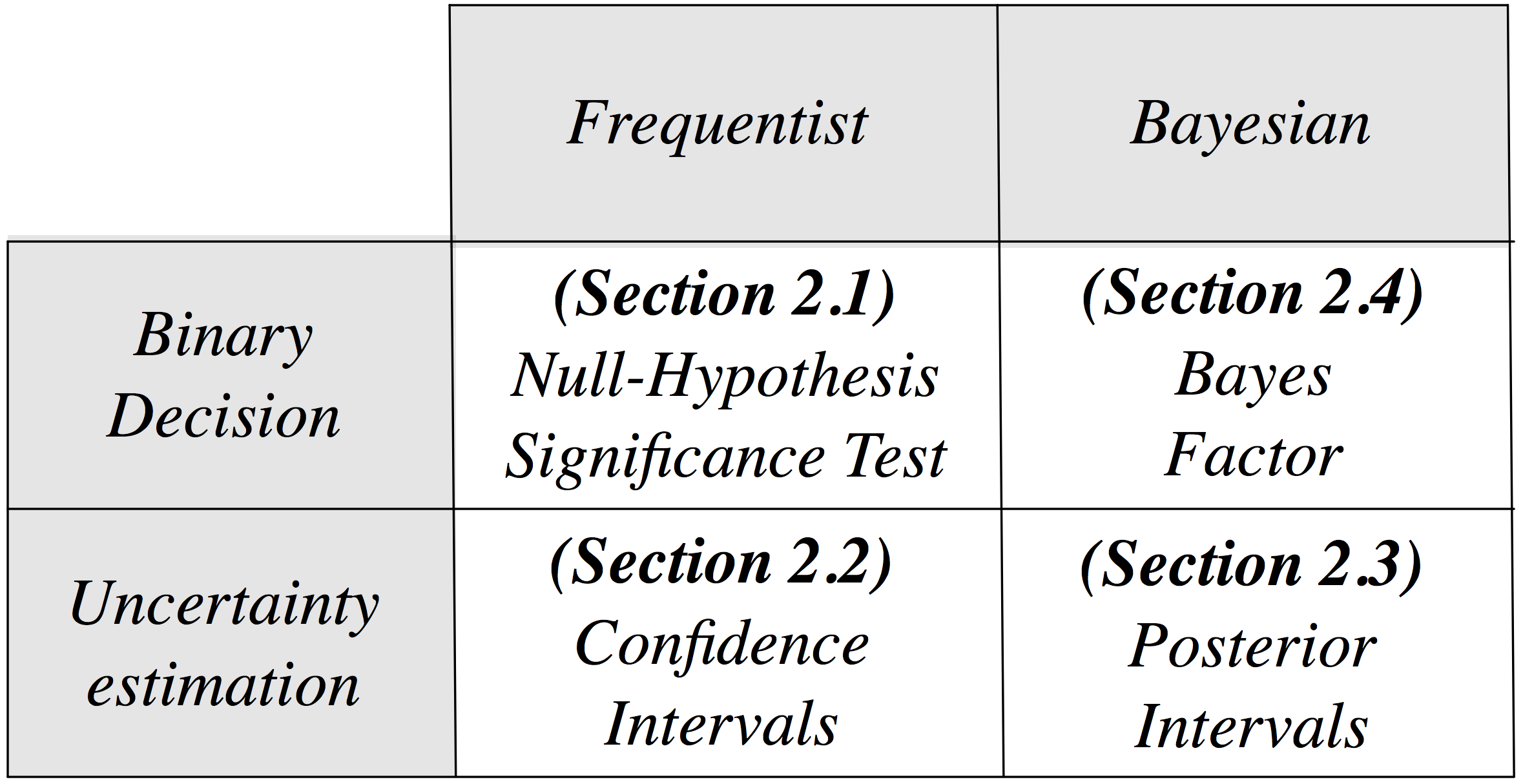}
    % \vspace{-0.1cm}
    \caption{
        Various classes of methods for statistical assessment of hypotheses. 
    }
    \label{tab:classes-of-methods}
\end{table}

Table~\ref{tab:classes-of-methods} shows a categorization of statistical tools developed for the assessment of such hypotheses.
% Among them, first tool is based on $p$-values (\S\ref{subsec:nhst:pvalue}) which is the most common approach.
The two tools on the left are based on \emph{frequentist} statistics, while the ones on the right are based on \emph{Bayesian} inference~\cite{kruschke2018bayesian}. 
A complementary categorization of these tools is based on the nature of the results that they provide: the ones on the top encourage \emph{binary} decision making, 
% among hypotheses, 
% while some provide insights into the uncertainty of the hypotheses.  
while those on the bottom provide uncertainty around estimates. We discuss all four classes of tests in the following sub-sections.

% the frequentist hypothesis testing, and then deep-dive into Bayesian hypothesis assessment. 
% Before we go into the details of Bayesian hypothesis testing, let us briefly review frequentist hypothesis testing. 

% \todo{
% somewhere we have to clarify the terminology, possibly by referring to Figure 2: 
%  - p-value + Bayes Factor --> significance test 
%  - the rest: summarizing the uncertainty. 
% }

\subsection{Null-Hypothesis Significance Testing}
\label{subsec:nhst:pvalue}
% \paragraph{frequentist hypothesis testing.}
% In  the Neyman-Pearson paradigm of 
In frequentist hypothesis testing, there is an asymmetric relationship between two hypotheses. The hypothesis formulated to be rejected is usually called the \emph{null-hypothesis} \hn. For instance, in our example \hn $:\theta_1 = \theta_2$. A decision procedure is devised by which, depending on $\boldsymbol{y}$, the null-hypothesis will either be \emph{rejected} in favor of \ha, or the test will stay \emph{undecided}. 
% or accepted. 

% difference of the two algorithms: $\delta(X) \triangleq \measure{A, \dataset} / \measure{B, \dataset}$; 
% and we define hypothesis testing problem: 
% $$
% \begin{cases}
% \emph{\hn}: & \delta(X) \geq \delta_0 \\ 
% \emph{\ha}: & \delta(X) < \delta_0 
% \end{cases}
% $$

% The probability that a test statistic that is as extreme or more extreme than the actual value could have arisen by chance, given the null-hypothesis. 

A key notion here is $p$-value, the probability, under the null-hypothesis \hn, of observing an outcome at least equal to or \emph{extreme} than the empirical observations $\boldsymbol{y}$. 
To apply this notion 
% the test 
on a set of observations $\boldsymbol{y}$, one has to define a function that maps $\boldsymbol{y}$ to a numerical value. 
% We will test statistics with $\delta(\boldsymbol{y})$ based on the performances of the two systems. 
% For instance, in the \emph{paired t-test} \cite{}, the average of performance differences is used, i.e., $\delta(\boldsymbol{y}) = \frac{1}{n}\sum_{i=1}^{n}(\measure{A, x_i} - \measure{B, x_i})$. 
% \todo{make it more accurate; refer to an existing test: for instance, in X test, it is defined as ... }
% $obs=\measure{A, \boldsymbol{x}} , \measure{B, \boldsymbol{x}}$
% $P(\delta(X)>=\delta(obs)|H_0)$
This function is called the \emph{test statistic} $\delta(.)$
% Notice that 
 and it formalizes the interpretation of \emph{extremeness}. Concretely, $p$-value is defined as, 
\begin{equation}
\label{equation:pvalue}
\prob{\delta(Y) \geq \delta(\boldsymbol{y}) | \emph{\hn} }
\end{equation}
In this notation, $Y$ is a random variable over possible observations and $\delta(\boldsymbol{y})$ is the empirically observed value of the test statistic.

% \todo{a good place to refer to other tests; talk about what `hitchhikers` says, etc. }
% \daniel{
% A small p-value you can conclude that your sample is so incompatible with the null-hypothesis that you can reject the null for the entire population;  although it does not necessarily mean that an effect is practically important. 
% A large (non-significant) p-value implies that the data could easily have been observed under the null-hypothesis. 
% }
% { \color{red}
% A large $p$-value implies that the data could easily have been observed under the null-hypothesis.
% However, lower values of $p-$values indicates that the observations are unlikely to obey the null-hypothesis (and thus, the null-hypothesis should be rejected). 
% }
A large $p$-value implies that the data could easily have been observed under the null-hypothesis.
Therefore, a lower $p$-value is used as evidence towards rejecting the null-hypothesis. 
% }

% The uncertainty of the estimated parameter value is represented a pre-define \emph{the significance level} $\alpha$, a fixed small constant. 

% Only if $p-$value is less than $\alpha$ (or, $\prob{\delta(Y) \geq \delta(\boldsymbol{y}) | \emph{\hn} } < \alpha$) then $\delta(y)$ is sufficiently unlikely and the null-hypothesis is rejected.
% The smaller the significance level $\alpha$, the less probable it is that the null-hypothesis holds. 

% An example use of $p$-value in our running example is provided below: 
% We provide an example of a using $p$-values in the context of our running example:  
\begin{example}[Assessment of \textbf{\emph{H1}}]
\label{exm:p-value}
We form a null-hypothesis using the accuracy of the two systems (Table~\ref{tab:table_output_dataset}) using  a one-sided $z$-test
%% NEW April 14:
\footnote{The choice of this test is based on an implicit assumption that two events corresponding to answering two distinct questions, are independent with identical probability, i.e., equal to the inherent accuracy of the system. Hence, the number of correct answers follows a binomial distribution. Since, the total number of questions is large, i.e., $2376$ in ARC-easy, this distribution can be approximated with a normal distribution. It is possible to use other tests with less restrictive assumptions (see \citet{dror2018hitchhiker}), but for the sake of simplicity we use this test to illustrate core ideas of ``p-value'' analysis.}
%% :NEW April 14
with
% for differences of two groups:
% comparing proportions of ... 
% (under the framework commonly referred to as one-sided z-test for comparing proportions of two groups): 
$\delta(\boldsymbol{y}) \triangleq   (1/n) \sum_{i=1}^{n} \left[ \measure{{S}_1, x_i} - \measure{{S}_2, x_i} \right].$ We formulate a null-hypothesis against the claim of ${S}_1$ having strictly better accuracy than ${S}_2$.
This results in a $p$-value of $0.0037$ (details in \S\ref{sec:details:ex1}) and can be interpreted as the following: \emph{if the systems have inherently identical accuracy values, the probability of observing a superiority at least as extreme as our observations} is $0.0037$. For a significance level of $0.05$ (picked before the test) this $p$-value is small enough to reject the null-hypothesis. 
\end{example}

This family of the tests is thus far the most widely used tool in NLP research. 
Each variant of this test is based on some assumptions about the distribution of the observations, under the null-hypothesis, and an appropriate definition of the test statistics $\delta(.)$. 
Since a complete exposition of such tests is outside the scope of this work, 
we encourage interested readers to refer to 
the existing reviews, such as \newcite{dror2018hitchhiker}.  
% \todo{Jordan: this paragraph seems mostly redundant}
% refer to the existing reviews of such techniques in NLP~\newcite{dror2018hitchhiker}. 
% and \newcite{genericFreq} in general. 
% For instance, for our running example, one can use binomial exact test~\cite{howell2012statistical} testing if each group has a certain ratio. We compare the proportions based on normal z-test. Take two systems: Reading Strategies and Multi-Task BERT (ensemble). We test the null-hypothesis against two cases: two-sided alternative and  Multi-Task BERT (ensemble) having strictly better accuracy than Reading Strategies. These two tests give $p-$values of $0.0074$ and $0.0037$ respectively. \todo{Do we need the formulas}
% The latter value can be interpreted as the following: if two systems have identical inherit\footnote{As opposed to only in a sample dataset.} accuracies, the probability of observing a superiority at least as extreme as (1721 vs 1637 out of 2376 question instances) is $0.0037$.

\subsection{Confidence Intervals}
\label{subsec:confidence:interval}
% \todo{Jordan: it's weird having CIs as a subsection between null-H and Bayes, since CI and null-H are not in opposition in the way null-H and Bayes are.}
% \todo{we need more accurate def here and our running example}
% \todo{OR we can dial this down and only talk in informal (at the same time ecouraging) wording?}
% In order to decide whether \hn\ should be rejected or not, one has to pre-define \emph{the significance level} $t$, a fixed threshold value. 

% the significance level is the area under the distribution curve bounded on the lower end by the actual value of the test statistic. The significance level is the probability that a test statistic that is as extreme or more extreme than the actual value could have arisen by chance, given the null-hypothesis. 
% Thus, the lower the significance level, the more likely it is that the two systems are significantly different. 

% - \emph{Sampling distributions} are at the core of p values and confidence intervals, 
% - For example, a traditional t-test involves computing a p value, and if p<.05 then the null-hypothesis is rejected. 

% \begin{figure}
%     \centering
%     \includegraphics[scale=0.5]{figures/frequentist2.png}
%     \caption{Caption}
%     \label{fig:my_label}
% \end{figure}

% \paragraph{Confidence Intervals.}
Confidence Intervals (CIs) are used to express the uncertainty of estimated parameters. 
% The uncertainty of estimated parameters can be expressed via Confidence Intervals (CIs). 
% \todo{The confidence interval then is a statement about the likelihood that the interval we obtain actually has the true parameter value.}
In particular, the $95$\% CI is the range of values for parameter $\boldsymbol{\theta}$ 
% such that the corresponding $p$-value is greater than or equal to $0.05$: 
such that the corresponding test based on $p$-value is not rejected: 
\begin{equation}
    \label{eq:ci_definition}
    \prob{\delta(Y) \geq \delta(\boldsymbol{y}) | \emph{\hn}(\boldsymbol{\theta} ) } \geq 0.05.
\end{equation}
In other words, the confidence interval merely asks which values of the parameter $\boldsymbol{\theta}$ could be used, before the test is rejected. 

\begin{example}[Assessment of \textbf{\emph{H2}}]
\label{exm:CI}
% \small
Consider the same setting as in Example \ref{exm:p-value}. 
According to Table 1, the estimated value of the accuracy differences (maximum-likelihood estimates) is 
$\theta_1 - \theta_2 = 0.035$.  
A 95\% CI of this quantity provides a range of values that are not rejected under the corresponding null-hypothesis. 
In particular, 
% a z-test provides 
a 95\% CI gives $\theta_1 - \theta_2 \in [0.0136, 0.057]$ (details in \S\ref{sec:details:ex2}). 
The blue bar in Figure~\ref{fig:HDI} (right) shows the corresponding CI. 
Notice that the conclusion of Example 1 is compatible with this CI; the null-hypothesis $\theta_1=\theta_2$ which got rejected is not included in the CI. 
% as we saw the hypothesis $\theta_1=\theta_2$ incurs a very low $p$-value and thus gets rejected.  

% The horizontal bar marks the 95\% CI, which is the range of parameter values that would not be rejected. In other words, any value for the parameter ?? outside the CI would be rejected by $p <0.05$.
\end{example}
% \erfan{ 
%     Consider the same setting as in Example 1. We saw that the hypothesis $\theta_1-\theta_2=0$ incure very low p-value and thus rejected. The hypotheses with $\theta_1-\theta_2\in[0.0136, 0.057]$ do not get rejected at 95 percent level. 
% }

\subsection{Posterior Intervals}
\label{subsec:posterior:interval}
Bayesian methods focus on prior and posterior distributions of 
$\boldsymbol{\theta}$. 
% the parameters of interest. 
Recall that in a typical NLP experiment, these parameters can be, e.g., the \emph{actual} mean or standard deviation for the performance of a system, as its inherent and unobserved property. 

% In Bayesian hypothesis testing, there can be more than two hypotheses under consideration, and they do not necessarily stand in an asymmetric relationship. 
% Rather, Bayesian hypothesis testing works just like any other type of Bayesian inference. 
In Bayesian inference frameworks, a priori assumptions and beliefs are encoded in the form of a \emph{prior} distribution $\mathbb{P}(\boldsymbol{\theta})$ on parameters of the model.\footnote{We use $\mathbb{P}(x)$ in its most general form, to denote the Probability Mass Function for discrete variables and the Probability Density Function for continuous variables. 
% For example, if $X$ is a discrete random variable then $\mathbb{P}(x)\triangleq \mathbb{P}(X=x)$.
} 
In other words, a prior distribution describes the common belief about the parameters of the model. It also implies a distribution over possible observations.
%%%%%%%%%%%%%%%%%55
% \erfan{Following the discussion in the email, I am adding this discussion here:}
For assessing hypotheses {\textbf{\textit{H3}}} and {\textbf{\textit{H4}}} in our running example, we will simply use the uniform prior, i.e., the inherent accuracy is uniformly distributed over $[0,1]$.  This corresponds to having no prior belief about how high or low the inherent accuracy of a typical QA system may be.

In general, the choice of this prior can be viewed as a compromise between the beliefs of the analyzer and those of the audience. The above uniform prior, which is equivalent to the Beta(1,1) distribution, is completely non-committal and thus best suited for a broad audience who has no reason to believe an inherent accuracy of 0.8 is more likely than 0.3.  For a moderately informed audience that already believes the inherent accuracy is likely to be widely distributed but centered around 0.67, the analyzer may use a Beta(3,1.5) prior to evaluate a hypothesis.  Similarly, for an audience that already believes the inherent accuracy to be highly peaked around 0.75, the analyzer may want to use a Beta(9,3) prior.
% \erfan{End of new text}
%%%%%%%%%%%%%%%%%%%%%%%%%555555
% Formally, one incorporates $\boldsymbol{\theta}$ in a hierarchical model for the distribution of observations $\mathbb{P}(\boldsymbol{y})$.
% { \color{red}
Formally, one incorporates $\boldsymbol{\theta}$ in a hierarchical model in the form of a \emph{likelihood function} $\prob{\boldsymbol{y}|\boldsymbol{\theta}}$. This explicitly models the underlying process that connects the latent parameters to the observations. 
% }
% \daniel{
% The Bayesian approach is characterized by the explicit use of probability distributions to describe the underlying process that results in the observations, commonly referred to as \emph{likelihood} distribution: $\prob{\boldsymbol{y}|\boldsymbol{\theta}}$
% }
% \todo{I think we should expand this by talking about likelihood; we can also include a figure to schematize the hierarchical model.}
Consequently, a \emph{posterior} distribution is inferred using the Bayes rule and conditioned on the  observations: 
    $\prob{\boldsymbol{\theta} | \boldsymbol{y}}=
        \frac{
            \prob{\boldsymbol{y}|\boldsymbol{\theta}}
            \prob{\boldsymbol{\theta}}
        }
        {\prob{\boldsymbol{y}}}.$

% \begin{equation}
%     \label{equation:posterior}
%     \prob{\boldsymbol{\theta} | \boldsymbol{y}}=
%         \frac{
%             \prob{\boldsymbol{y}|\boldsymbol{\theta}}
%             \prob{\boldsymbol{\theta}}
%         }
%         {\prob{\boldsymbol{y}}} .
% \end{equation}

% The posterior distribution is a combined summary of 
% of what 
% the data and prior information and tell us about which values of $\boldsymbol{\theta}$ are more likely.
% The posterior distribution is a combined summary of 
% the data and prior information and it highlights the likely values of $\boldsymbol{\theta}$.
The posterior distribution is a combined summary of the data and prior information, about likely values of $\boldsymbol{\theta}$.
The mode of the posterior (maximum a posteriori)
can be seen as an estimate for 
% each of components of 
$\boldsymbol{\theta}$. 
Additionally, the posterior can be used to describe 
% an 
% a detailed 
% expression of 
the \emph{uncertainty} around the mode.

While the posterior distribution can be analytically calculated for simple models, it is not so straightforward for general models. Fortunately, recent advances in hardware, Markov Chain Monte Carlo (MCMC) techniques \cite{metropolis1953equation,gamerman2006markov}, and probabilistic programming\footnote{$\mathsf{Pymc3}$ (in Python) and $\mathsf{JAGS}$ \& $\mathsf{STAN}$ (in R) are among the commonly-used packages for this purpose.} allow sufficiently-accurate numerical approximations of posteriors.

\begin{figure*}[ht]
\centering
\includegraphics[width=0.99\textwidth,trim=0cm 1.6cm 0cm 3.1cm, clip=false]{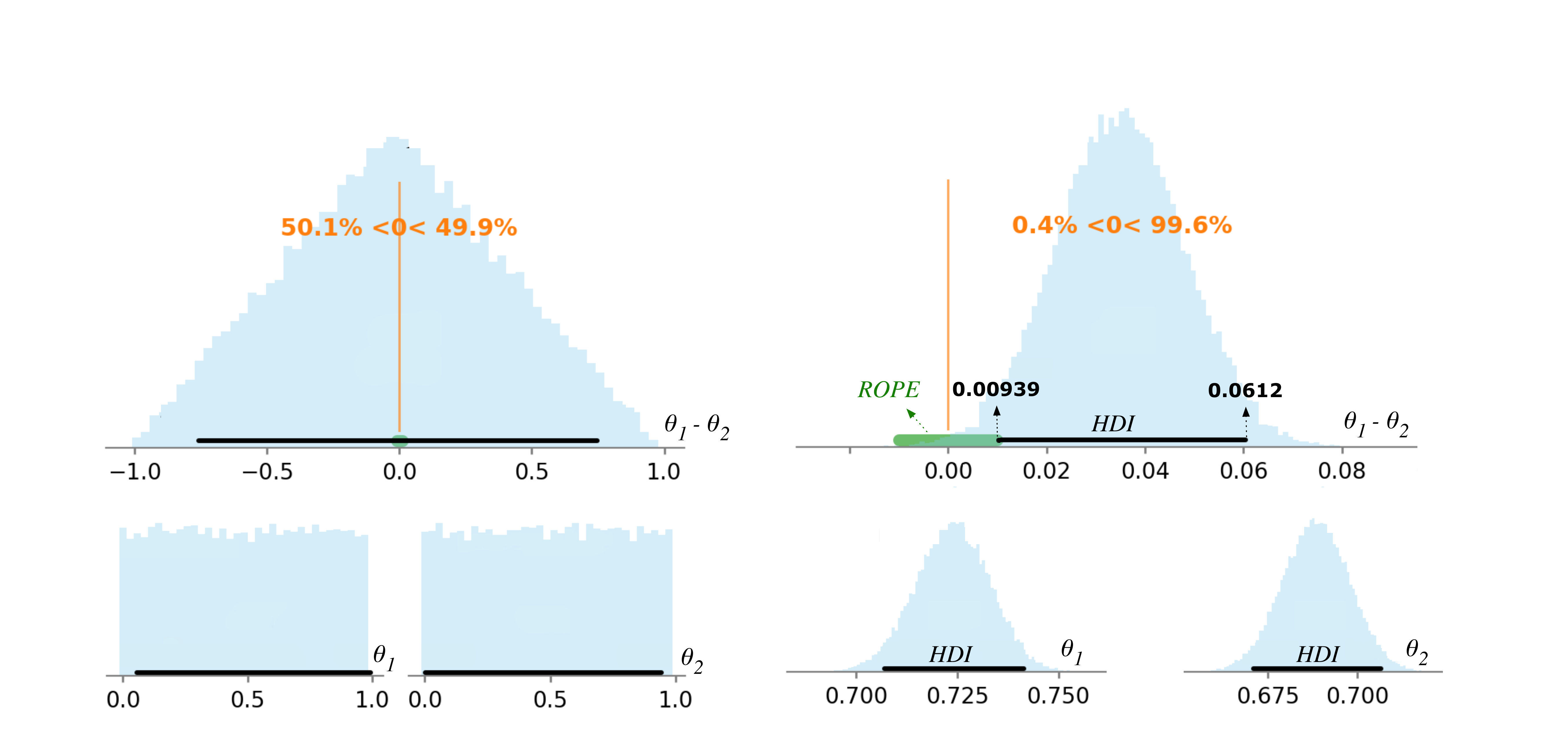}
\caption{
% \small
Left: Prior distributions of two systems (bottom row) and their difference (top row). 
Right: 
Posterior distributions of two systems (bottom row) and their difference (top row) after observing the performances on ARC-easy dataset. 
% Note that the posterior HDI estimate is $(0.00939, 0.0612)$. 
Note the posterior HDI estimate, $(0.00939, 0.0612)$. 
Here we assume at least one percent 
% to be the threshold for the difference of accuracies 
accuracy difference 
to be considered practically different. 
Hence, we indicate the interval $(-0.01, 0.01)$ as \textit{\color{DarkGreen}ROPE} (\S\ref{subsec:posterior:interval}.) 
% 1: which overlaps with $95\%$ HDI $(0.00939, 0.0612)$
% 2: and $95\%$ HDI $(0.00939, 0.0612)$. 
% 3: HDI is estimated as $(0.00939, 0.0612)$. 
% The interval is highlighted in red and the middle of the interval, indicating where the accuracies are exactly equal, is indicated with a green line in all figures.
}
\label{fig:HDI}
\end{figure*}

% \begin{figure*}[h]
% \centering
% \includegraphics[width=0.47\textwidth]{figures/BinomialQACompare_prior_theta_big.png}
% \includegraphics[width=0.47\textwidth]{figures/BinomialQACompare_posterior_theta_big.png}
% \caption{
% \small
% Left: Prior distributions of two systems (bottom row) and their difference (top row). 
% Right: 
% Posterior distributions of two systems (bottom row) and their difference (top row) after observing the performances on ARC-easy dataset.
% Here we assume at least one percent 
% % to be the threshold for the difference of accuracies 
% accuracy difference 
% to be considered practically different. 
% Hence, we indicate the interval $(-0.01, 0.01)$ to be the ROPE (introduced in \S\ref{subsec:posterior:interval}.) The interval is highlighted in red and the middle of the interval, indicating where the accuracies are exactly equal, is indicated with a green line in all figures.\footnote{The figure could be reproduced via the accompanying software.}
% }
% \label{fig:HDI2}
% \end{figure*}

% \paragraph{Uncertainty summarization via posterior.}
% \paragraph{Testing by formulating uncertainty.}
% 1.Hypothesis assessment based on posterior.
% 2.Expression of uncertainty via summarizing the posterior.
% 3. Uncertainty summarization via posterior
% 4. From posterior to Expression of uncertainty
One way to summarize the uncertainty around the point estimate of parameters is by marking the span of values that cover $\alpha$\% of the most-credible density in the posterior distribution (e.g., $\alpha = 95$\%). 
% The interval that a high percentage of probability density for posterior is condensed. 
% {
% \color{blue}
% These are called the \emph{Highest Density Intervals} (HDIs). 
% Notice that Confidence Intervals are commonly mistaken to represent the same concept as HDIs. That's why some refer to HDI as Bayesian Confidence Intervals~\cite{oliphant2006bayesian}
This is called \emph{Highest Density Intervals} (HDIs) or Bayesian Confidence Intervals~\cite{oliphant2006bayesian} (not to be confused with CI, in \S2.2). 

Recall that a hypothesis $H$ is a condition on $\boldsymbol{\theta}$ (see Figure~\ref{fig:notation}).
% This means that
Therefore, given the posterior $\prob{\boldsymbol{\theta}|\boldsymbol{y}}$, one can calculate the probability of $H$, as a probabilistic event, conditioned on $\boldsymbol{y}$: $\prob{H|\boldsymbol{y}}$. 

For example in an unpaired $t$-test, \hn\ is the event that the means of two groups are equal.
Bayesian statisticians usually relax this strict equality $\theta_1=\theta_2$ and instead evaluate the credibility of $|\theta_1-\theta_2|<\varepsilon$ for some small value of $\varepsilon$. 
The intuition is that when $\theta_1$ and $\theta_2$ are close enough they are \emph{practically} equivalent.
%Formally, \hn\ can be extended to refer to as the event that the absolute difference of means is less than a threshold. 
% This motivates statisticians to define
This motivates the definition of 
% This is captured by 
\emph{Region Of Practical Equivalence} (ROPE): An interval around zero with ``negligible'' radius.
The boundaries of ROPE depend on the application, the meaning of the parameters and its audience.
In our running example, a radius of one percent for ROPE implies that improvements less than $1$ percent are not considered notable. 
For a discussion on setting ROPE see~\newcite{kruschke2018rejecting}.

% Given these terms assessing a hypothesis $H$ is formulated as estimating $\prob{H|\boldsymbol{y}}$. This is trivial from the posterior $\prob{\theta|\boldsymbol{y}}$. Here are a few examples of hypotheses one can consider for $H$ in this formulation

These concepts give researchers the flexibility to define and assess a wide range of hypotheses. 
% Take the following different instances that can be of interest depending on the application:
For instance, we can address {\textbf{\emph{H3}}} (from Introduction) and its different variations that can be of interest depending on the application. The analysis of {\textbf{\emph{H3}}} is depicted in Figure~\ref{fig:HDI} and explained next.\footnote{Figure~\ref{fig:HDI} can be readily reproduced via the accompanying software, 
\hybayes.}

\begin{comment}
\begin{tcolorbox}[
                colback=LightBlue,colframe=blue!125!black,
                  left=0.5pt,
                  right=0.5pt,
                  top=0.5pt, 
                  bottom=0.5pt, 
                  boxrule=0.1mm
                  ]
\textit{
    \begin{enumerate}[(2.a),topsep=0.1pt,itemsep=0.1ex,partopsep=0.1ex,parsep=0.1ex,leftmargin=*]
        \item The mean performance of $S_1$ is strictly higher than the mean performance of $S_2$.
        \item The difference of means of performances of two systems is within ROPE.
        \item The margin of superiority of mean performance of $S_1$  vs $S_2$ is greater than ROPE.
    \end{enumerate}
}
\end{tcolorbox}
\end{comment}

% Daniel: commented this out, since it doesn't seem to add to much here, without enough context 
\begin{comment}
Notice that the probabilities of the above hypotheses can be extracted from posterior distribution.
\end{comment}

\begin{example}[Assessment of \textbf{\emph{H3}}]
\label{exm:HDI}
Recall the setting from previous examples. 
% about the systems for the question answering task introduced in \S\ref{sec:preliminaries}. 
The left panel of Figure~\ref{fig:HDI} shows the prior on the latent accuracy of the systems and their differences
% Notice that, here, our prior belief about the accuracies of two systems is a uniform distribution.
% over $[0, 1]$. 
% In appendix we discuss other choices for prior too.
% For the details of the hierarchical model used here and a discussion on the choice of the prior refer to \S\ref{sec:details:ex3}. 
% Here we consider the experiment where two systems are tested on ``ARC-easy'' dataset. 
(further details on the hierarchical model in \S\ref{sec:details:ex3}.)
We then obtain the posterior distribution (Figure~\ref{fig:HDI}, right), in this case via numerical methods). 

Notice that one can read the following conclusion: with probability $0.996$, the hypothesis \emph{\textbf{H3}} 
% (with $x=0$) 
(with {\color{red}a margin} of 0\%) 
holds true. 
% Notice that one can read the following conclusion: \emph{With probability $0.996$ the accuracy of $S_1$ is higher than $S_2$.} This corresponds to the statement~\emph{\textbf{H3}} with $x=0$. 
As explained in \S\ref{subsec:practical_significance}, this statement does \underline{not} imply any difference with a notable margin. In fact, the posterior in Figure~\ref{fig:HDI} implies that this experiment is \underline{not} sufficient to claim the following: with probability at least $0.95$,
hypothesis \emph{\textbf{H3}} 
% (with $x=1$) 
(with {\color{red}a margin} of 1\%) 
holds true. 
This is the case since {\color{DarkGreen}ROPE} $(−0.01,0.01)$ overlaps with $95\%$ HDI $(0.00939, 0.0612).$ 
% (as indicated in the figure.)  
% The researcher might proceed with performing another experiment with another dataset. Figure~\ref{fig:post2} shows the posterior given both observations from the performances on easy and challenge datasets.
% Notice that in this case, HDI is completely higher than two percent superiority of the accuracy of A over B. This means that one can make the following statement: With probability $0.95$, system A's accuracy is two percent higher than that of B's. In this case, it seems acceptable to informally state this claim as: system A practically significantly outperforms system B.
\end{example}

\subsection{Bayes Factor}
\label{subsec:bayes:factor}
% A common approach within Bayesian framework is called Bayesian hypothesis testing, where the notion of \emph{Bayes Factor} is introduced.
% A common tool within Bayesian framework is the notion of \emph{Bayes Factor} (sometimes referred to as \emph{Bayesian hypothesis testing}).
A common tool among Bayesian frameworks is the notion of \emph{Bayes Factor}.\footnote{\emph{``Bayesian Hypothesis Testing''} usually refers to the arguments based on \emph{``Bayes Factor.''} However, as shown in \S\ref{subsec:posterior:interval}, 
there are other Bayesian approaches for assessing hypotheses.}
% it is not the only Bayesian approach for assessing hypotheses.}
% This measure, intuitively, compares the degree to which the density shifts from one prior to posterior for each hypothesis: 
Intuitively, 
it
% Bayes Factor 
compares how the observations $\boldsymbol{y}$ shift the credibility from prior to posterior of the two competing hypothesis: 

% (1) While \emph{Bayesian Hypothesis Testing} usually refers to Bayes Factor, here take a more general definition by viewing any approach that uses summary of posterior information (including \S\ref{subsec:posterior:interval}. 
% (2) While \emph{Bayesian Hypothesis Testing} usually refers to Bayes Factor, as we shown in \S\ref{subsec:posterior:interval}, it is not the only Bayesian approach to assessing hypotheses. 
% (2') Note that \emph{Bayesian Hypothesis Testing} usually refers to the arguments based on Bayes Factor. However, as shown in \S\ref{subsec:posterior:interval} it is not the only Bayesian approach to assessing hypotheses. 
% (3) Note that only the arguments based on Bayes Factor usually refereed to as \emph{Bayesian Hypothesis Testing}. We take a more general view of Bayesian hypothesis assessment (including both \S\ref{subsec:bayes:factor} and \S\ref{subsec:posterior:interval}). 
% Bayes Factor is the ratio of the marginal likelihood of two competing hypotheses: 
$$
\mathit{BF}_{01} = 
% \frac{ \prob{ \boldsymbol{y} | \text{\hn}} }{ \prob{ \boldsymbol{y} | \text{\ha}} } =  
\frac{\prob{\text{\hn} | \boldsymbol{y}}}{\prob{\text{\ha} | \boldsymbol{y}}} \Bigg/ \frac{\prob{\text{\hn}}}{\prob{\text{\ha}}}
$$

% In the case of a null-hypothesis vs alternate hypothesis, Bayes Factor can be approximated via Savage-Dickey method \cite{dickey1971weighted}:

% $$
% \mathit{BF}_{01} \approx \frac{
%                     \text{Probability density of the ROPE in Prior}
%                 }{
%                     \text{Probability density of the ROPE in Posterior}
%                 }
% $$

If the $\mathit{BF}_{01}$ equals to 1 then the data provide equal support for the two hypotheses and there is no reason to change our a priori opinion about the relative likelihood of the two hypotheses. 
A smaller Bayes Factor is an indication of rejecting the  null-hypothesis $H_0$. 
If it is greater than 1 then there is support for the null-hypothesis and 
% we should increase the odds in favor of \hn. 
we should infer that the odds are in favor of \hn.

Notice that the symmetric nature of Bayes Factor allows all the three outcomes of ``accept'', ``reject'', and ``undecided,'' as opposed to the definition of $p$-value that cannot accept a hypothesis.

\begin{example}[Assessment of \textbf{\emph{H4}}]
\label{exam:bayesFactor}
Here we want to assess the null-hypothesis \hn : $|\theta_1-\theta_2|<0.01$ against \ha:  $|\theta_1-\theta_2|\geq 0.01$ ($x=0.01$). 
% Using a bit of calculations on the posterior, once can show that:  
Substituting posterior and prior values, one obtains: 
% {\color{red}$\mathit{BF}_{01}=\frac{0.02760125
% }{0.9800325} \Bigg/ \frac{0.0199675
% }{0.97239875
% } = 1.382308752$.}
$$\mathit{BF}_{01}=\frac{0.027
}{0.980} \Bigg/ \frac{0.019
}{0.972
} = 1.382$$.
This value is very close to $1$ which means that this observation does not change our prior belief about the two systems difference. 
% the probabilities this specific hypothesis has not changed much after observing the data. 
%However for a smaller ROPE this value happens to be XX. In this case, it is implied that the observed value highly support the alternative hypothesis.
%For our running example, Bayes Factor happens to be XX and XX for the case of only considering observations on easy dataset or challenge as well.
\end{example}

\section{Comparisons}
% \subsection{Fallacies Involving $p$-value}
% \section{Misinterpretations \& Malpractices of $p$-values}
% \paragraph{$p$-values malpractices.}
% \paragraph{$p$-values misinterpretations.}
\label{sec:comparisons}

Many aspects influence the choice of an approach to assess significance of hypotheses. 
% Here we briefly go over a few of the well-known issues and compare the aforementioned techniques.
% This section compares different  techniques in terms of various well-known issues. 
This section provides a comparative summary, with details in Appendix~\ref{appendix:sec:comparisons} and an overall summary in Table~\ref{tab:method-comparison}. 

% For a more detailed comparison, please refer to the appendix \S\ref{appendix:sec:comparisons}. 

\begin{table*}[t]
    \setlength{\doublerulesep}{\arrayrulewidth}
    \centering
    \small
    \scalebox{0.81}{
    \begin{tabular}{|C{2.1cm}|c|C{1.8cm}|C{2.0cm}|C{1.8cm}|C{2.5cm}|C{2.6cm}|C{1.8cm}|}
        \hline \hline
        \T Method & Paradigm & \T Ease of interpretation ($1=$easy) (\S\ref{subsec:main_text:misinterpretation}) & Encourages binary-thinking (\ref{subsec:comparison:measures:of:uncertainty}) & Depends on stopping intention (\ref{subsec:main_text:sensitivity_sample_size}) & Dependence on prior (\ref{subsec:main_text:priorSesitivity}) & Decision rule &  \# of papers using this test in ACL'18 \\
        \hline \hline
        \T (\S\ref{subsec:nhst:pvalue}) $p$-value & frequentist & 3 & Yes & Yes & No & \T Acceptable $p$-value & 73 \\ 
        \hline
        \T (\S\ref{subsec:confidence:interval}) CI & frequentist & 4 & No & Yes & No & \T Acceptable confidence margin &  6\\ 
        \hline
        \T (\S\ref{subsec:posterior:interval}) HDI & Bayesian  & 1 & No &  No & Not sensitive but takes it into account & \T HDI relative to ROPE & 0\\ 
        \hline
        \T (\S\ref{subsec:bayes:factor}) BF & Bayesian  & 2 & Yes & No & Highly sensitive & \T Acceptable BF & 0\\ 
        \hline \hline
    \end{tabular}
    }
    % \vspace{-0.1cm}
    \caption{A comparison of different statistical methods for evaluating the credibility of a hypothesis given a set of observations.
    The total number of published papers in at the ACL-2018 conference is 439. 
    }
    \label{tab:method-comparison}
\end{table*}

\subsection{Susceptibility to Misinterpretation}
% \paragraph{Ambiguity in interpretation.}
% \paragraph{Ambiguity in interpretation.}
% \paragraph{Easiness to misinterpret.}
% \paragraph{Susceptibility to misinterpret.}
% \paragraph{Misinterpretation.}
\label{subsec:main_text:misinterpretation}

The complexity of interpreting significance tests combined with insufficient reporting could result in ambiguous or misleading conclusions.
This ambiguity can not only confuse authors but also cause confusion among readers of the papers.

While $p$-values (\S\ref{subsec:nhst:pvalue}) are the most common approach, they are inherently complex, which makes them easier to misinterpret (see examples in \S\ref{subsec:misinterpretation}).
Interpretation of confidence intervals (\S\ref{subsec:confidence:interval}) can also be challenging since it is an extension of $p$-value~\cite{hoekstra2014robust}. Approaches that provide measures of uncertainty directly in the hypothesis space (like the ones in \S\ref{subsec:posterior:interval}) are often more natural choices for reporting the results of experiments~\cite{kruschke2018bayesian}. 
% Overall, our view on ``ease of interpretation'' of the approaches is summarized in Table~\ref{tab:method-comparison}. 

\subsection{Measures of Certainty}
\label{subsec:comparison:measures:of:uncertainty}
A key difference is that not all methods studied here provide a measure of uncertainty over the hypothesis space.
For instance, $p$-values (\S\ref{subsec:nhst:pvalue}) do \emph{not} provide probability estimates on two systems being different (or equal)~\cite{goodman2008dirty}.
On the contrary, they encourage \emph{binary} thinking~\cite{Gelman13theproblem}, that is, confidently concluding that one system is better than another, without taking into account the extent of the difference between the systems.
% Thus, they which often does not say anything about the relative merits of two hypotheses. 
CIs (\S\ref{subsec:confidence:interval}) provide a range of values for the target parameter. However, this range also does not have any \emph{probabilistic} interpretation in the hypothesis space~\cite{du2009confidence}. 
On the other hand, posterior intervals (\S\ref{subsec:posterior:interval}) generally provide a useful summary as they capture probabilistic estimates of the correctness of the hypothesis.

\subsection{Dependence on Stopping Intention}
% \paragraph{Sensitivity to the sample size.}
\label{subsec:main_text:sensitivity_sample_size}
% \label{subsec:dependence_stopping_intention}
% The process based on which the data is sampled could affect the results of a test. In particular, 

The process by which samples in the test are collected can affect the outcome of a test. For instance, the sample size $n$ (whether it is determined before the process of gathering information begins, or is a random variable itself) can change the result. 
Once observations are recorded, this distinction is usually ignored. Hence, the testing algorithms that do not depend on the distribution of $n$ are more desirable. Unfortunately, the definition of $p$-value (\S\ref{subsec:nhst:pvalue}) depends on the distribution of $n$. For instance, \citet[\S11.1]{kruschke2010bayesian} provides examples where this subtlety can change the outcome of a test, even when the final set of observations is identical.

\subsection{Sensitivity to the Choice of Prior}
\label{subsec:main_text:priorSesitivity}
% \paragraph{Sensitivity to the choice of prior.}
The choice of the prior can change the outcome of Bayesian approaches (\S\ref{subsec:posterior:interval} \& \S\ref{subsec:bayes:factor}). 
% It is a well-known issue that 
Decisions of Bayes Factor (\S\ref{subsec:bayes:factor}) are known to be sensitive to the choice of prior, while posterior estimates (\S\ref{subsec:posterior:interval}) are less so. For further discussion, see \ref{subsec:mitigateBFSens} or refer to discussions by 
\citet{sinharay2002sensitivity,liu2008bayes} or \citet{dienes2008understanding}.

\section{Current Trends and Malpractices}
% \section{Comparison, Misinterpretations, and Fallacies}
% \subsection{Fallacies Involving $p$-value}
% \section{Misinterpretations \& Malpractices of $p$-values}
% \paragraph{$p$-values malpractices.}
% \paragraph{$p$-values misinterpretations.}
% \section{Usage Patterns in the Field}
% \section{Common Practices in the Field}
\label{sec:common_practices}

This section highlights common practices relevant to the our target approaches. 
To better understand the common practices or misinterpretations in the field, we conducted a survey. 
% on the understanding and interpretation of hypothesis testing techniques. 
We shared the survey among $\sim$450 NLP researchers (randomly selected from ACL'18 Proceedings)
% who have published in the past few years 
% and 40 individuals decided to participate in our survey. 
from which 55 individuals filled out the survey. 
While similar surveys have been performed in other fields~\cite{windish2007medicine}, this is the first in the NLP community, to the best of our knowledge.
Here we review the main highlights (see Appendix for more details and charts).

    \paragraph{Interpreting $p$-values.}
    While the majority of the participants have a self-claimed ability to interpret $p$-values (Figure~\ref{fig:know_pvalues}), many choose its imprecise interpretation \emph{``The probability of the observation this extreme happening due to pure chance''} (the popular choice) vs.\ a more precise statement \emph{``Conditioned on the null hypothesis, the probability of the observation this extreme happening.''} (see \emph{Q1} \& \emph{Q2} in Appendix~\ref{sec:survey_details}.)
    
    \paragraph{The use of CIs.}
     Even though $95\%$ percent of the participants self-claimed the knowledge of CIs (Figure~\ref{fig:know_confidence_intervals}), it is rarely used in practice. 
    In an annotation done on ACL'18 papers by two of the authors, only 6 (out of 439) papers were found to use CIs. 
    % To get a little sense of the popularity of the CIs in the community, two authors went over the ACL'18 papers in order to annotate those papers that use CIs. The authors found only 6 papers that use confidence intervals for their quantitative results in order to indicate some measure of uncertainty. 

\begin{figure}
    \centering
    \includegraphics[scale=0.52,trim=0.4cm 18.7cm 11.0cm 0cm, clip=false]{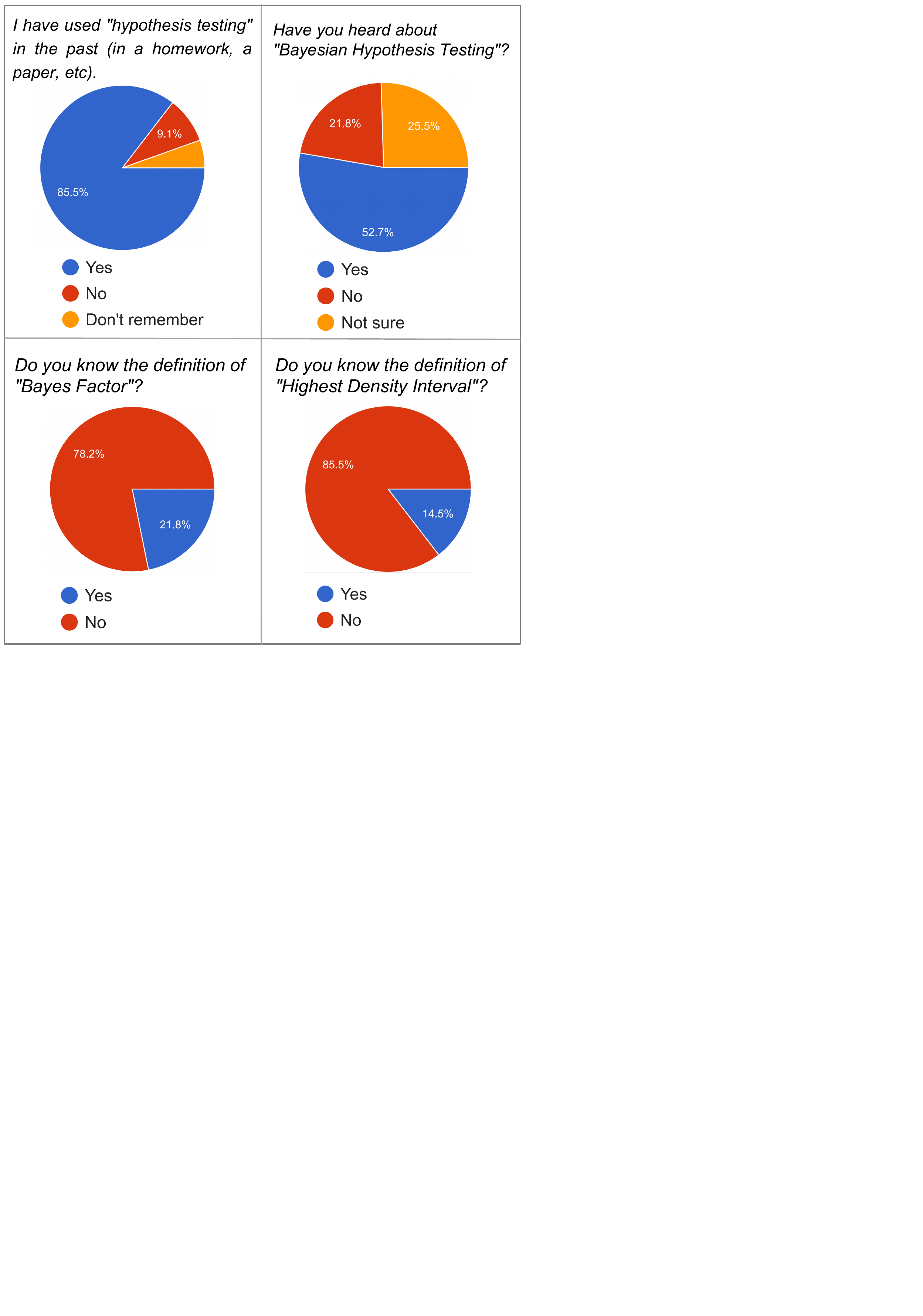}
    % \vspace{-0.1cm}
    \caption{Select results from our survey. 
    % of NLP researchers.
    }
    \label{fig:survey:combined:figure}
\end{figure}

    \paragraph{The use of Bayes Factors.}
     A majority  of the participants 
    %  ($57\%$; Figure~\ref{fig:heard_bayesian_hypothesis_testing}) 
     had ``heard'' about ``Bayesian Hypothesis Testing'' but did not know the definition of ``Bayes Factor'' (Figure~\ref{fig:survey:combined:figure}). %  (Figure~\ref{fig:bayes_factor}).  
    HDIs (discussed in \S\ref{subsec:posterior:interval}) 
    were the least known.  
    We did not find any papers in ACL'18 that use Bayesian tools.

    % Comparing the answers to Q3.1 and Q3.2 reveals that the participants are more familiar with the less accurate interpretation of $p-$value: ``The probability of the observation this extreme happening due to pure chance'' vs more precise statement ``Conditioned on the null hypothesis, the probability of the observation this extreme happening.'' 
    
    \paragraph{The use of ``significan*''.}
    A notable portion of NLP papers express their findings by using the term ``significant'' (e.g., \emph{``our approach significantly improves over X.''}) 
    Almost all ACL'18 papers use the term ``significant''\footnote{Or other variants ``significantly'', ``significance'', etc.} somewhere. 
    Unfortunately, there is no single universal interpretation of such phrases across readers. 
    % In some cases this could refer to statistical significance test, and in some other cases it could be a claim about the margin of improvement (and these two are different claims). 
    In our survey, we observe that when participants read ``X significantly improves Y'' in the abstract of a hypothetical paper: 
% ``... yields statistically significant better results in terms of all ROUGE metrics with p < 0.01.`` --> ambiguous sentence. 
    % Q3.3: Writers and readers in NLP community do not agree on the meaning of significance claims:
    \begin{enumerate}[topsep=0.5pt,itemsep=0.5ex,partopsep=0.5ex,parsep=0.5ex,leftmargin=*]
        \item 
        % {
        % \color{red}
        About $82\%$ expect the claim to be backed by ``hypothesis testing''; however, only $57\%$ expect notable empirical improvement (see \emph{Q3} in Appendix~\ref{sec:survey_details});
        % }
        % Even though $82\%$ of the participants expected the term ``significance'' claim to be backed by hypothesis testing, only $57\%$ interpreted it as having notable empirical margin. 
        % meaning of ``statistical significance'', i.e., a notable margin in the observations, to be presented 
        
       \item 
       About $35\%$ expect the paper to test ``practical significance'', which is not generally assessed by popular tests (see \S\ref{subsec:practical_significance});
    %   to be shown in the paper, which is not a result of significance tests.
    %   $35\%$ of participants expected the meaning of ``practical significance'' to be shown in the paper, which is not a result of significance tests.
       \item
       A few also expect a theoretical argument.  % to back a significance test.
    \end{enumerate}

\paragraph{Recent trends.}
% \subsection{Comparison of Assessment Methods}
% \label{sec:comparisons}

Table~\ref{tab:method-comparison} provides a summary of the techniques studied here. We make two key observations: (i) many papers don't use any hypothesis assessment method and would benefit from one; (ii) from the final column, $p$-value based techniques clearly dominate the field, a clear disregard to the advantages that the bottom two alternatives offer.

\section{Recommended Practices}
\label{sec:recommended_practices}

Having discussed common issues, we provide a collection of recommendations (in addition to the prior recommendations, such as by \newcite{dror2018hitchhiker}).
% These items are designed to be complementary to
% Refer to the literature that guides choosing the correct test, e.g., 
% the recommendations mentioned in \newcite{dror2018hitchhiker}.  

% for the field to follow for any of the choices in \S\ref{sec:preliminaries}. 
% Importantly, each of the tools in \S\ref{sec:preliminaries} provide different types of information and cover different aspects of the hypothesis assessment. That is why it is up to a researcher to choose the method they deem appropriate based on their specific question, e.g., how high the probability of their desired hypothesis being true is or how low the probability of observing such extreme difference in performances is if in reality there was no inherent difference. 
% \todo{expand this: a paragraph encouraging users to define their question first and choose the right technique according to their question.}
% In any case, here we provide some concrete recommendations to encourage the researchers to take a mindful approach especially by effective understanding, reporting and interpretation.

% of confidence intervals and uncertainty of the predictions~\cite{cumming2014new}. 
% \todo{Explain stopping intention in the issues or prelims}
% \begin{enumerate}[topsep=0.1pt,itemsep=0.1ex,partopsep=0.1ex,parsep=0.1ex,leftmargin=*]
    % \paragraph{Defining the question.}
The first step is to \underline{define your goal}. Each of the tools in \S\ref{sec:assessment} provides a distinct set of information. Therefore, one needs to formalize a hypothesis and consequently the question you intend to answer by assessing this hypothesis. Here are four representative questions, one for each method:

    \begin{tcolorbox}[
                colback=blue!2!white,colframe=blue!125!black,
                  left=0.5pt,
                  right=0.5pt,
                  top=0.5pt, 
                  bottom=0.5pt, 
                  boxrule=0.1mm
                  ]
        \emph{
        \small
        \begin{enumerate}[topsep=0.1pt,itemsep=0.1ex,partopsep=0.1ex,parsep=0.1ex,leftmargin=*]
            \item {\color{red} Assuming that the null-hypothesis is true}, is it likely  to witness observations this extreme? (\S\ref{subsec:nhst:pvalue})
            \item How much my null-hypothesis can {\color{red}deviate from the mean} of the observations until a $p$-value argument rejects it. (\S\ref{subsec:confidence:interval})
            \item Having observed the observations, {\color{red}how probable} is my claimed hypothesis?(\S\ref{subsec:posterior:interval})
            \item By observing the data how much {\color{red}do the odds increase} in favor of the hypothesis?(\S\ref{subsec:bayes:factor})
        \end{enumerate}
        }
    \end{tcolorbox}

    \begin{table*}[t]
        \centering
        \includegraphics[scale=0.78,trim=0.5cm 2.35cm 0cm 1.6cm, clip=false]{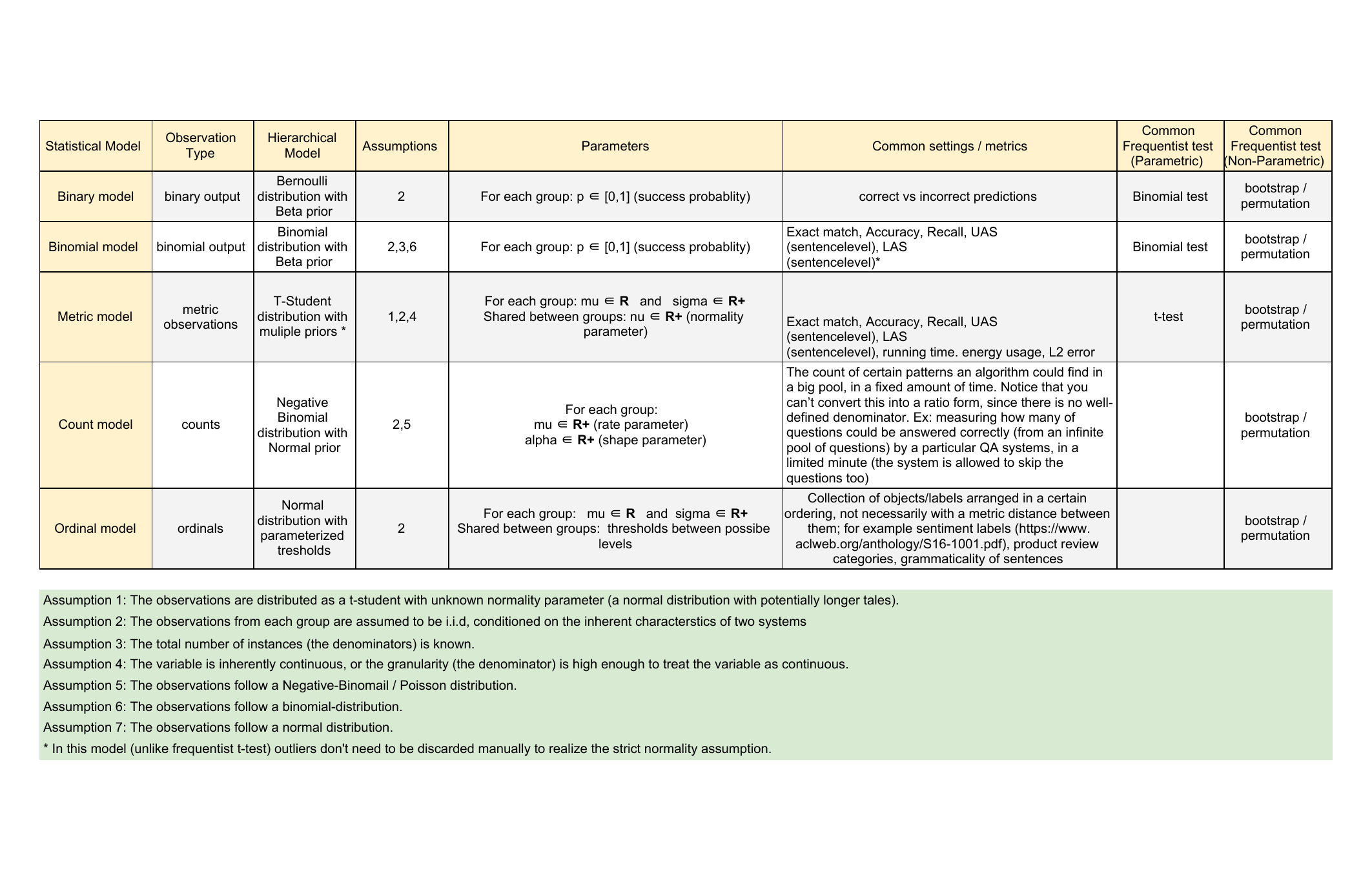}
        % \vspace{-0.2cm}
        \caption{Select models supported by our package \hybayes at the time of this publication.}
        \label{tab:HyBayes}
    \end{table*}

    \noindent
    If you decide to use \textbf{frequentist tests}: 
    % \paragraph{The frequentist way.}
    % \item For a frequentist approach: 
    \begin{itemize}[topsep=0.5ex,itemsep=0.5ex,partopsep=0.1ex,parsep=0.1ex,leftmargin=*]
        % \item Refer to the literature that guides choosing the correct test, e.g., \newcite{dror2018hitchhiker}.
        \item Check if your setting is compatible with the assumptions of the test. In particular, investigate if the meaning of null-hypothesis and  sampling distribution 
        % significance level, the stopping intention, etc, 
        % matche the real conditions in which the experiments have been done.
        match the experimental setting. 
        \item Include a summary of the above investigation. 
         Justify unresolved assumption mismatches.
        \item Statements reporting $p$-value and confidence interval must be precise enough so that the results are not misinterpreted (see \S\ref{subsec:main_text:misinterpretation}). 
        % It should be clear that these arguments only talk about significance of the observations, conditioned on the null-hypothesis (rather than probability of the hypothesis being false). 
        \item The term ``significant'' should be used with caution and clear purpose to avoid misinterpretations (see \S\ref{sec:common_practices}). One way to achieve this is by using adjectives  ``statistical'' or ``practical'' before any (possibly inflected) usage of ``significance.''  
        % Following the issues in \S\ref{subsec:practical_significance} and corresponding results in our survey, use the adjectives either ``statistical'' or ``practical'' before any, possibly in an inflected form, usage of the word ''significance.''  
        \item Often times, the desired conclusion is a notable margin in the superiority of one system over another (see \S\ref{sec:comparisons}). In such cases, a pointwise $p$-value argument is \emph{not} sufficient; a confidence interval analysis is needed. If CI is inapplicable for some reason, this should be mentioned.
    \end{itemize}

    \noindent If you decide to use \textbf{Bayesian approaches}: 
    % \paragraph{The Bayesian way.}
    % \item For a Bayesian approach: 
    \begin{itemize}[topsep=0.5ex,itemsep=0.5ex,partopsep=0.1ex,parsep=0.1ex,leftmargin=*]
        \item Since Bayesian tests are less 
        % commonly 
        known, it is better to provide a short motivation for the usage. 
        % (vs frequentist tests). 
        % in the publications.
        % \item Aquatint yourself with flexible ways of expressing a hierarchical model. { \color{blue} You can craft a customized model or utilize one of the commonly used specifications ; e.g. the software accompanying this work or the R scripts, publicly available in \newcite{kruschke2010bayesian}. 
        % }
        \item  Familiarize yourself with packages that help you decide  a hierarchical model, e.g., the software provided here. If necessary, customize these models for your specific problem. 
        
        \item Be clear about your hierarchical model, including model parameters and priors. In most cases, these choices should be justified (see \S\ref{subsec:posterior:interval}.)
        % Also meaning and intuition about the main parameters of interest, e.g., difference of means of two populations, need to be highlighted.
        
        % \item The most important part of a Bayesian Analysis the posterior distribution of the main parameters or hypotheses, e.g. 
        % difference of means of two inherit accuracy values. 
        % the difference between two inherent accuracy values. 
        
        % Along with a plot of this distribution, include details on Central Tendencies, e.g., mean and mode, ROPE, HDI, etc. See Figure \ref{fig:post1} for an example.
        
        \item Comment on the certainty (or the lack of) of your inference in terms of HDI and ROPE: (I) is HDI completely inside ROPE, (II) they are completely disjoint, (III) HDI contains values both inside and outside ROPE (see \S\ref{subsec:posterior:interval}.)
        
        % \item { \color{blue} For reproducibility and further investigations, make the MCMC traces, convergence plots, effective sample size of key parameters publicly available. Note that the software accompanying the paper provides all of these.  }

        \item For reproducibility, include further details about your test:  MCMC traces, convergence plots, etc. (Our \hybayes package provides all of this.)
        
        % \item Use Bayes Factor when a relative comparison of two models/hypotheses is your intention.
        
        \item Be wary that Bayes Factor is highly sensitive to the choice of prior (see \S\ref{subsec:main_text:priorSesitivity}). See Appendix \S\ref{subsec:mitigateBFSens} for possible ways to mitigate this. 
        % { \color{blue} If there are a few potential priors that all can describe your vague intuition of the a priori knowledge, then a quick hack is to use a small subset of the observations to get a slightly-informed prior (see \S\ref{sec:informedPrior} for an example) before running main Bayes Factor analysis.} Also, reporting the results from several choices of priors can mitigate the concerns about sensitivity to prior. 
        % Notice that one can avoid such concerns by only using the approach in \S\ref{subsec:posterior:interval}. 
    \end{itemize}  
% \end{enumerate}
% \todo{ Jordan: I think section 2 can be cut down significantly, because while it's good to give some summary, someone who wants to implement this stuff will have to consult a statistics text anyway. With the space you save, you could give more practical concrete NLP examples, involving typical tables with P-R-F1 scores and so on. }
% }

\subsection{Package \hybayes}

We provide an accompanying package, \hybayes, to facilitate comparing systems using the two Bayesian hypothesis assessment approaches discussed earlier: 
% , based on given observations. The package mainly supports two Bayesian approaches for statistical assessment:
(a) posterior probabilities and (b) Bayes Factors. (Several packages are already available for frequentist assessments.)
% Both of these approaches can be used to claim superiority of one algorithm over another from different aspects.
%

Table~\ref{tab:HyBayes} summarizes common settings in which \hybayes can be employed\footnote{These settings are available at the time of this publication, with more options likely to be added in the future.} in NLP research, including typical use cases, underlying data assumptions, recommended hierarchical model, metrics (accuracy, exact match, etc.), and frequentist tests generally used in these cases. These settings cover several typical assumptions on observed NLP data. However, if a user has specific information on observations or can capitalize on other assumptions, we recommend adding a custom model, which can be done relatively easily.

%%%%%%%%%%%%%%%%%%%%%%%%%%
\section{Conclusion}

Using well-founded mechanisms for assessing the validity of hypotheses is crucial for any field that relies on empirical work.
Our survey indicates that the NLP community is not fully utilizing scientific methods geared towards such assessment, with only a relatively small number of papers using such methods, and most of them relying on $p$-value. 

% Such tests encourage binary thinking about presence or absence of effects and such an approach is not necessarily meaningful in the real world; The acceptance of an experiment as valid research result should not be based on a binary decision rule~\cite{serlin1993rational,serlin1985rationality}. 
% To compensate with such issues, other fields like statistics and psychology (andy, hopefully, our field as well) are shifting the emphasis away from null-hypothesis significance testing to ``estimation based on effect sizes, confidence intervals and metal-analysis''~\cite{cumming2014new}. In this light, we reviewed Bayesian approaches to hypothesis testing and to estimation with confidence or credible intervals. Moreover, we release a software to help the practitioners in NLP use Bayesian tests for their models. 

Our goal was to review different alternatives, especially a few often ignored in NLP. We surfaced various issues and potential dangers of careless use and interpretations of different approaches. 
% Our goal is to provide a unifying view of the pitfalls and best practices, and equip NLP researchers with Bayesian hypothesis assessment approaches as an important tool in their toolkit.
We do not recommend a particular approach. Every technique has its own weaknesses. Hence, a researcher should pick the right approach according to their needs and intentions, with a proper understanding of the techniques. Incorrect use of any technique can result in misleading conclusions. 

% mindless use of statistics and acknowledges that mindless use is bad in both freq. and bayes. 
% That's why he says it is necessary to examine the hierarchical model everytime we use and do not assume a well known test has the best model ever we need.

% However, I think if the community accepts the bayesian approach as "easy 
% to interpret" approach they would gradually invest more time in proper usage.

% A scientific conclusion or should be based on whether or not p-value is significant.

% \todo{
% Add a paragraph on this: 
% "Bayesian statistics, in short, can’t save us from bad science."
% \url{https://www.nytimes.com/2014/09/30/science/the-odds-continually-updated.html?_r=0}

% Or 
% mindless use of statistics and acknowledges that mindless use is bad in both freq. and bayes. 
% That's why he says it is necessary to examine the hierarchical model everytime we use and do not assume a well known test has the best model ever we need.

% However, I think if the community accepts the bayesian approach as "easy 
% to interpret" approach they would gradually invest more time in proper usage.
% }

% \section{Future Work}
% more specific hierarchical models.
% Meta analysis: to get most certain conclusion across different publications for each NLU task.
% Power analysis. 

We contribute a new toolkit, \hybayes, to make it easy for NLP practitioners to use Bayesian assessment in their efforts. 
We hope that this work provides a \emph{complementary} picture of hypothesis assessment techniques for the field and encourages more rigorous reporting trends.
% treatment of such techniques. 

\subsection*{Acknowledgments}
{
% \small    %% font seems too small and widely spaced
\fontsize{10}{10}\selectfont
The authors would like to thank Rotem Dror, Jordan Kodner, and John Kruschke for invaluable feedback on an early version of this draft. 
% We also thank the 50+ anonymous participants in our NLP researcher survey.
This work was partly supported by a gift from the Allen Institute for AI and by DARPA contracts FA8750-19-2-1004 and FA8750-19-2-0201.
% and by contract FA8750-13-2-0008 with the US Defense Advanced Research Projects Agency (DARPA). 
% The views expressed are those of the authors and do not reflect the official policy or position of the U.S. Government.
}

\bibliographystyle{acl_natbib}
\bibliography{ref}

\clearpage

\appendix

% \twocolumn

\section{Additional Details: Examples}
\label{sec:supplemental}

\subsection{More Details: Example~\ref{exm:p-value}}
\label{sec:details:ex1}
Here we use a one-sided z-test to compare $s_1=1721$ out of $2376$ vs $s_2=1637$ out of $2376$.
We start with calculating the z-score:
\begin{gather}
s_1 = 1721, n_1 = 2376\\
s_2 = 1637, n_2 = 2376\\
p_1 = s_1 / n_1 = 0.72432\\
p_2 =  s_2 / n_2 = 0.68897\\
\tilde{p} = (s_1 + s_2) / (n_1 + n_2) = 0.70664\\
z = \frac { p_1 - p_2} {\sqrt{\tilde{p}(1-\tilde{p})((1/n_1)+(1/n_2))}} =2.676 
\end{gather}
Then we can read one-sided tail probability from a z-score table corresponding to  $2.676$ as $0.00372$.

\subsection{More Details: Example~\ref{exm:CI}}
\label{sec:details:ex2}

Let $\tilde{z}$ denote the z-score corresponding to 95\%, i.e.,  $\tilde{z} = 1.644853$. Also, let $s$ be denominator of the z-score formula above, i.e., $\sigma = \sqrt{\tilde{p}(1-\tilde{p})((1/n_1)+(1/n_2))}$. Then the confidence interval is calculated as follows:
\[ [p_1 - p_2 - \tilde{z} \sigma , p_1 - p_2 + \tilde{z} \sigma] =  [0.0136, 0.057].
\]

% Example 3 and 4 calculated using the accompanying package

\subsection{More Details: Example~\ref{exm:HDI}}
\label{sec:details:ex3}

\paragraph{Hierarchical model.}
In this analysis, the input consists of four non-negative positive integers $a_1,n_1,a_2,$ and $n_2$. The $i$th algorithm has answered $a_i$(out of $n_i$) questions correctly. 
% Here we assume that $n_1$ and $n_2$ are fixed. 
In our model, we assume that $a_i$ follows a binomial distribution with parameters $\theta_i$ and $n_i$. Note that this is mathematically the same as considering $n_i$ Bernoulli random variables with $a_i$ of them being success and $n_i-a_i$ being failure.

\begin{figure}[h]
    \centering
    \includegraphics[scale=0.35]{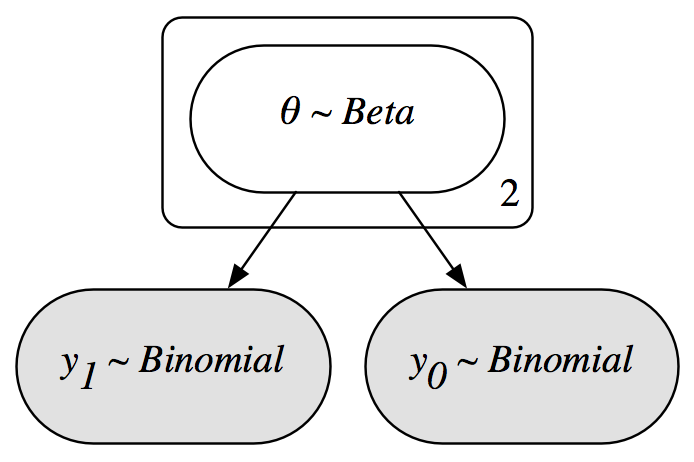}
    \caption{The hierarchical model in Example \ref{exm:HDI}\&\ref{exam:bayesFactor}. }
    \label{fig:hierarchical}
\end{figure}

At the higher level, $\theta_i$ is assumed to follow a uniform distribution in $[0, 1]$. In a later section below, we show how slightly different values for $Beta$ distribution can be used instead as a generalization. 
To run this analysis with \hybayes one can run the following command:

{
\small
% \begin{minted}{python}
\begin{verbatim}
python -m HyBayes \ 
       --config config_QABinomial.ini
\end{verbatim}
}
\noindent where {\small \texttt{config\_QABinomial.ini}} is a configuration file that defines a hierarchical model and the locations of the observations. This particular config file corresponds to the 2nd row of Table~\ref{tab:HyBayes}. 
For more information and examples, please refer to the manual of our released package (footnote 1). 

With this command, \hybayes would run a sampling-based Bayesian inference on the observed data, using the specified hierarchical model. The results should look like Fig.~\ref{fig:HDI}.

% The configuration file {\small \texttt{config\_QABinomial.ini}} can be accessed through this
% \hyperlink{https://github.com/allenai/HyBayes/blob/master/configs/config_QABinomial.ini}{link}
% . As indicated in the first section of this configuration file, the package needs the data files too. Those files are given in this  
% \hyperlink{https://github.com/allenai/HyBayes/tree/master/docs/example_analysis/real_data}{folder}
% to run the analysis. For more information refer to the \hyperlink{https://github.com/allenai/HyBayes/blob/master/docs/MANUAL.md}{extended manual} of the package.

% Here is { \textsc pymc3} code for specifying this model:
% {
% \small
% % \begin{minted}{python}
% \begin{verbatim}
% a = 1
% b = 1
% with pm.Model() as pymcModel:
%   theta = pm.Beta(
%     "theta", a, b,shape=2)
%   observations = []
%   for i in range(2):
%     observations.append(
%       pm.Binomial(f'y_{i}',
%         n=y[i],
%         p = theta[i],
%         observed = y[i][:, 1])
%     )
% \end{verbatim}
% }

\paragraph{Generalizations.}
It is possible to take previous observations in the literature into account and start with a non-uniform prior.
For example, setting $\alpha$ and $\beta$ to $\frac{1}{2}$ incorporates the idea that the performances are generally closer to $0$ or $1$, on the other hand $\alpha=\beta=2$ lowers the probability of such extremes.
Notice that, as long as $\theta_1$ and $\theta_2$ are assumed to follow the same distribution, the probability that $S_1$ is better than $S_2$ is $0.5$,  as expected from any fair prior.

\begin{figure}[ht]
\centering
\includegraphics[width=0.43\textwidth,trim=1cm 3cm 0cm 1.9cm]{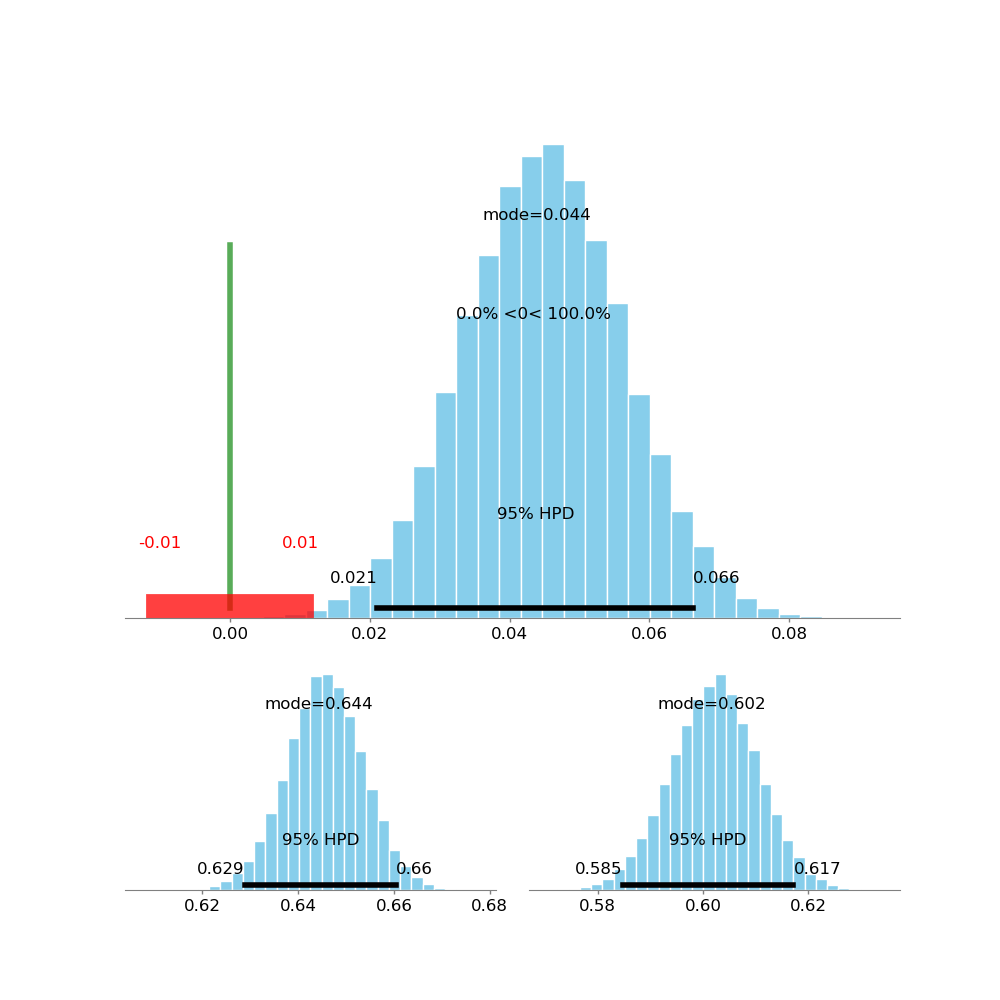}
\caption{Posterior distributions of two systems (bottom row) and their difference (top row) after observing the performances on both datasets.}
\label{fig:post2}
\end{figure}

\paragraph{New observations.}
The researcher might proceed with performing another experiment with another dataset. Fig.\ref{fig:post2} shows the posterior given both observations from the performances on the ``easy'' and ``challenge'' datasets.
Notice that in this case, HDI is completely higher than two percent superiority of the accuracy of $S_1$ over $S_2$. This means that one can make the following statement: With probability $0.95$, system $S_1$'s accuracy is two percent higher than that of $S_2$'s. In this case, it seems acceptable to informally state this claim as: system $S_1$ practically significantly outperforms system $S_2$.

\clearpage 

\onecolumn

\section{Additional Details About the Survey in Section~\ref{sec:common_practices}}
\label{sec:survey_details}

% In \S\ref{sec:common_practices}, we cite several results as the outcome of a survey of NLP researchers. 
% Here we provide further details on this survey (Fig.\ref{fig:distribution_of_degrees}-\ref{fig:q3})
Below we include three questions from the survey discussed in Section~\ref{sec:common_practices} that measure participants' interpretation of various terms or statistical tools. Next to each question, we show the distribution of the collected responses. 
In Q1 \& Q2, reveal that individuals have substantial difficulties in interpreting $p-$values. 
% In \emph{none} of the following questions, the most popular answer is the correct answer. 
In Q1, while (b) is the popular response, the correct answer is the much less popular option (c). 
In Q2, while (d) is the most popular response, the correct answer is (c). 

For Q3, as discussed in \S\ref{sec:common_practices}, researchers often find it difficult to interpret statements that contain ``significant''. Therefore, we must be extra cautious about how this loaded keyword is used.

% \FrameSep3pt

\begin{figure*}[ht]
    \centering
          \begin{subfigure}[b]{0.67\textwidth}
            \begin{framed}
            \footnotesize
            \noindent 
            \emph{
            \textbf{Q1:} An NLP paper shows a performance of 38\% for a classifier. They also show that adding a feature F improves the performance to 39\%. In this finding, the authors have stated that $p$-value $> 0.05$. This means:
            }
            \begin{enumerate}[(a),topsep=0.3ex,itemsep=0.3ex,partopsep=0.0ex,parsep=0.0ex,leftmargin=2.5ex]
                \item The probability is greater than 1 in 20 that an improvement would be found again, if the study were repeated.
                \item The probability is greater than 1 in 20 that an improvement this large could occur by chance alone.
                \item The probability is less than 1 in 20 that an improvement this large could occur by chance alone.
                \item The chance is 95\% that the study is correct.
                \item I don't know, because I don't know what $p$-value means.
                \item Don't have time. Skipping this question.
            \end{enumerate}
            \end{framed}
    \end{subfigure}
    \begin{subfigure}[b]{0.3\textwidth}
        \centering
        \includegraphics[scale=0.45,trim=0.23cm 0.45cm 0cm 0cm, clip=true]{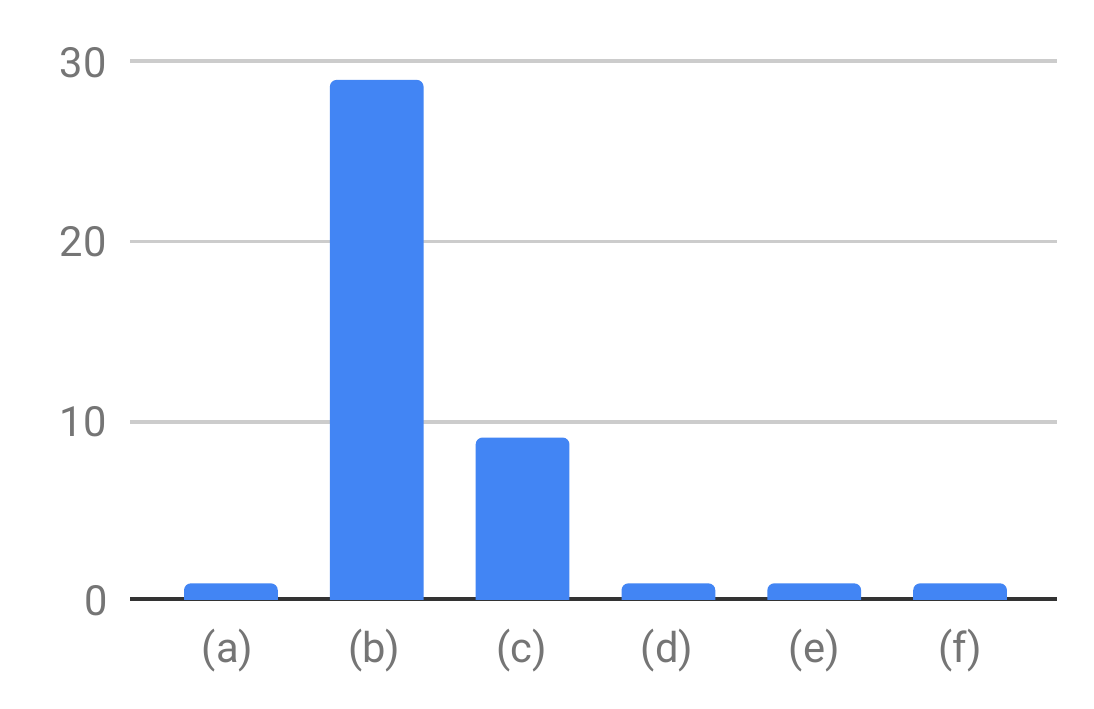}
        \caption{Distribution of responses to \emph{Q1}.}
        \label{fig:q1}
    \end{subfigure}
\end{figure*}

\begin{figure*}[ht]
    \centering
          \begin{subfigure}[b]{0.67\textwidth}
            \begin{framed}
            \footnotesize
            \noindent 
            \emph{
            \textbf{Q2:} An NLP paper shows a performance of 38\% for a classifier-1. They also show that adding a feature improves the performance to 45\% (call this classifier-2). The authors claim that this finding is "statistically significant" with a significance level of 0.01. Which of the following(s) make sense?
            }
            \begin{enumerate}[(a),topsep=0.3ex,itemsep=0.3ex,partopsep=0.0ex,parsep=0.0ex,leftmargin=2.5ex]
                \item With probability at least 0.01 classifier-2 will have better results than classifier-1, if we repeat this experiment.
                \item With probability 0.01 the observations are sampled from two equal classifiers.
                \item The probability of observing a margin ≥ 7\% is at most 0.01, assuming that the two classifiers inherently have the same performance.
                \item If we repeat the experiment, with a probability 99\% classifier-2 will have a higher performance than classifier-1.
                \item If we repeat the experiment, with a probability 99\% classifier-2 will have a performance margin of 7\% compared to classifier-1.
                \item I don't know, because I don't know how hypothesis testing is done.
                \item Don't have time. Skipping this question.
            \end{enumerate}
            \end{framed}
        %  \caption{$y=x$}
        %  \label{fig:y equals x}
    \end{subfigure}
    \begin{subfigure}[b]{0.3\textwidth}
        \centering
        \includegraphics[scale=0.45,trim=0.23cm 0.45cm 0cm 0cm, clip=true]{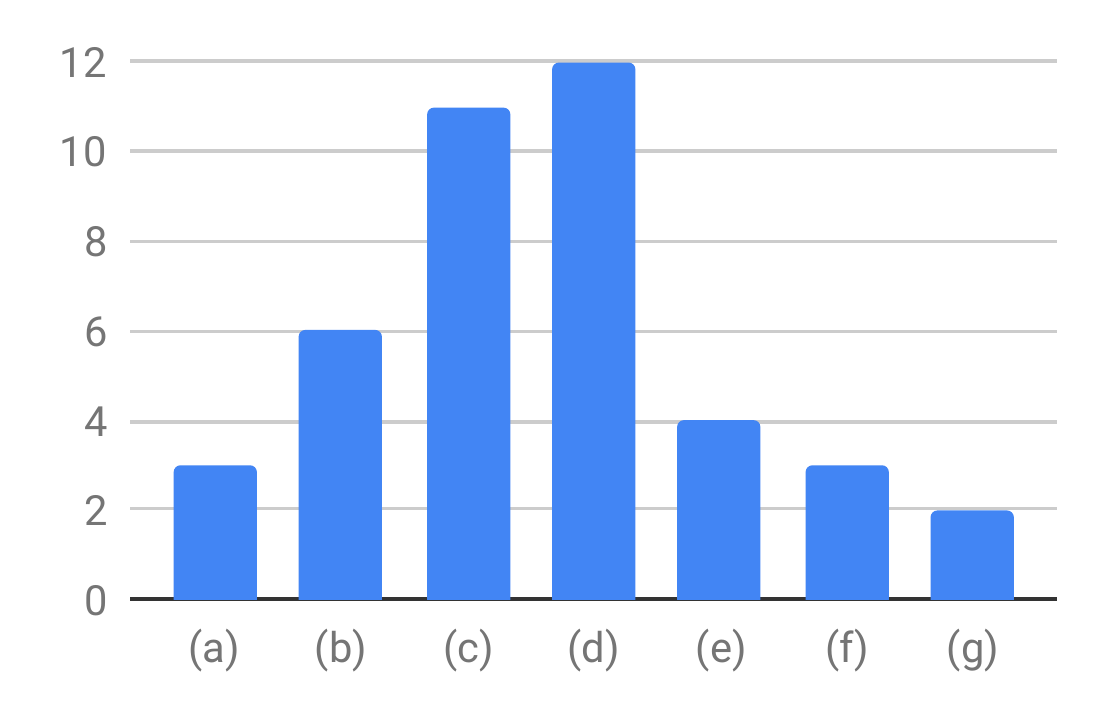}
        \caption{Distribution of responses to \emph{Q2}.}
        \label{fig:q2}
    \end{subfigure}
\end{figure*}

\begin{figure*}[ht]
    \centering
          \begin{subfigure}[b]{0.67\textwidth}
             \begin{framed}
                \footnotesize
                \noindent 
                \emph{
                \textbf{Q3:} An NLP paper presents system-1 and it compares it with a baseline system-2. In its "abstract" it writes: " ... system-1 significantly improves over system-2." What are the right way(s) to interpret this (select all that applies) 
                }
                \begin{enumerate}[(a),topsep=0.3ex,itemsep=0.3ex,partopsep=0.0ex,parsep=0.0ex,leftmargin=2.5ex]
                    \item It is expected that authors have performed some type of "hypothesis testing."
                    
                    \item It is expected that the authors have reported the performances of two systems on a dataset where system-1 has a higher performance than system-2 with a notable margin in the dataset.
                    
                    \item It is expected that the authors have reported experiments and concluded that system-1, inherently, has a higher perf. than system-2 with a notable margin.
                    
                    \item It is expected that the authors conclude through theoretical arguments that, system-1 has a higher inherent perf. than system-2 with a notable margin.
                    
                    \item Even though not expected, it is acceptable that the authors are concluding, through some theoretical arguments, that system-1 has a higher inherent performance than system-2 with a notable margin.
            
                    \item Too tedious. Skipping this question.
                \end{enumerate}
                \end{framed}
        %  \caption{$y=x$}
        %  \label{fig:y equals x}
    \end{subfigure}
    \begin{subfigure}[b]{0.3\textwidth}
         \centering
         \includegraphics[scale=0.45,trim=0.23cm 0.45cm 0cm 0cm, clip=true]{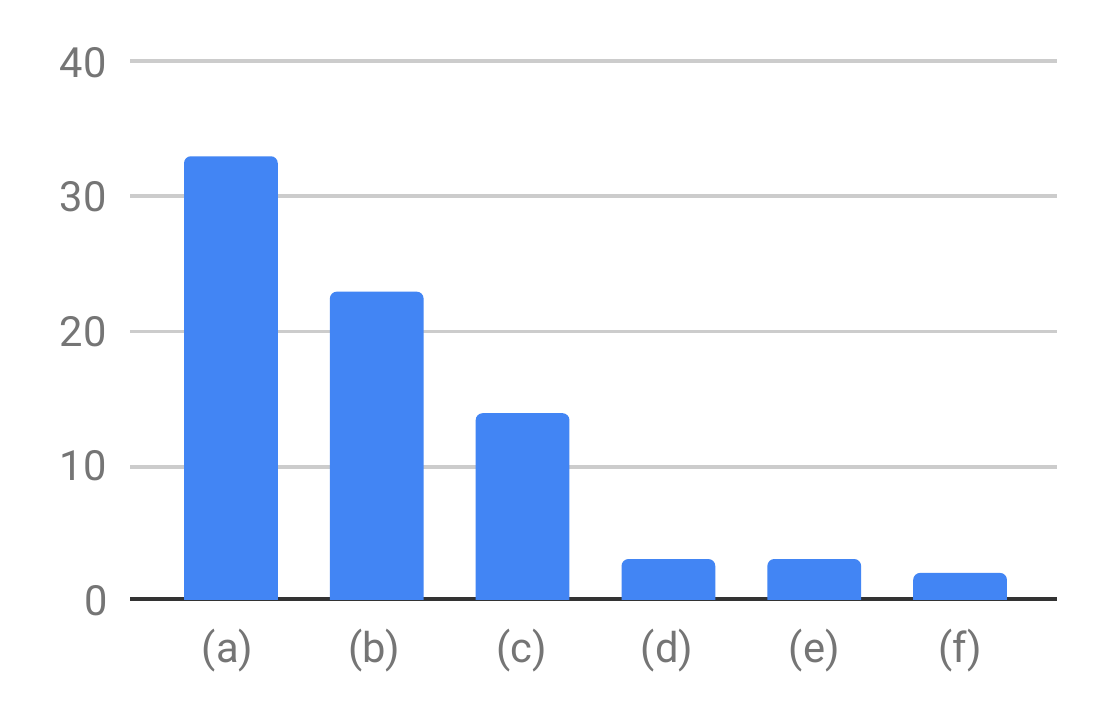}
        \caption{Distribution of responses to \emph{Q3}.}
         \label{fig:y equals x}
         \label{fig:q3}
     \end{subfigure}
\end{figure*}

\clearpage

Responses to the remaining questions are summarized in Figure~\ref{figure:results:of:survey}:

\begin{figure*}[ht]
    \centering
    \vspace{2ex}
    \begin{subfigure}[t]{0.49\textwidth}
        \centering
        \raisebox{-\height}{\includegraphics[scale=0.19,trim=0cm 0cm 0cm 0.5cm, clip=true]{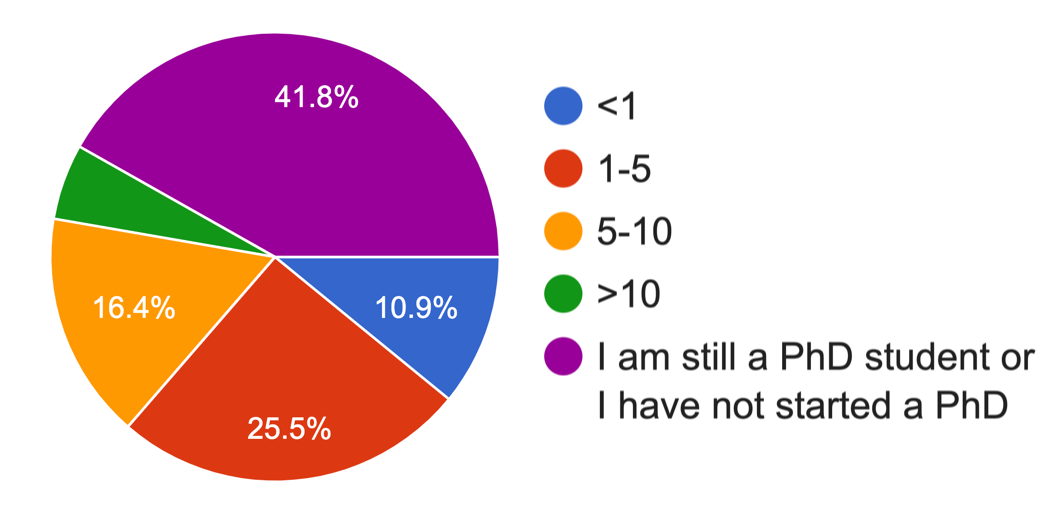}}
        \caption{Current positions of the participants in our survey.}
        \label{fig:distribution_of_degrees}
    \end{subfigure}
    \hfill
    \begin{subfigure}[t]{0.49\textwidth}
        \centering
        \raisebox{-\height}{\includegraphics[scale=0.47,trim=0.5cm 0.2cm 0cm 0cm, clip=true]{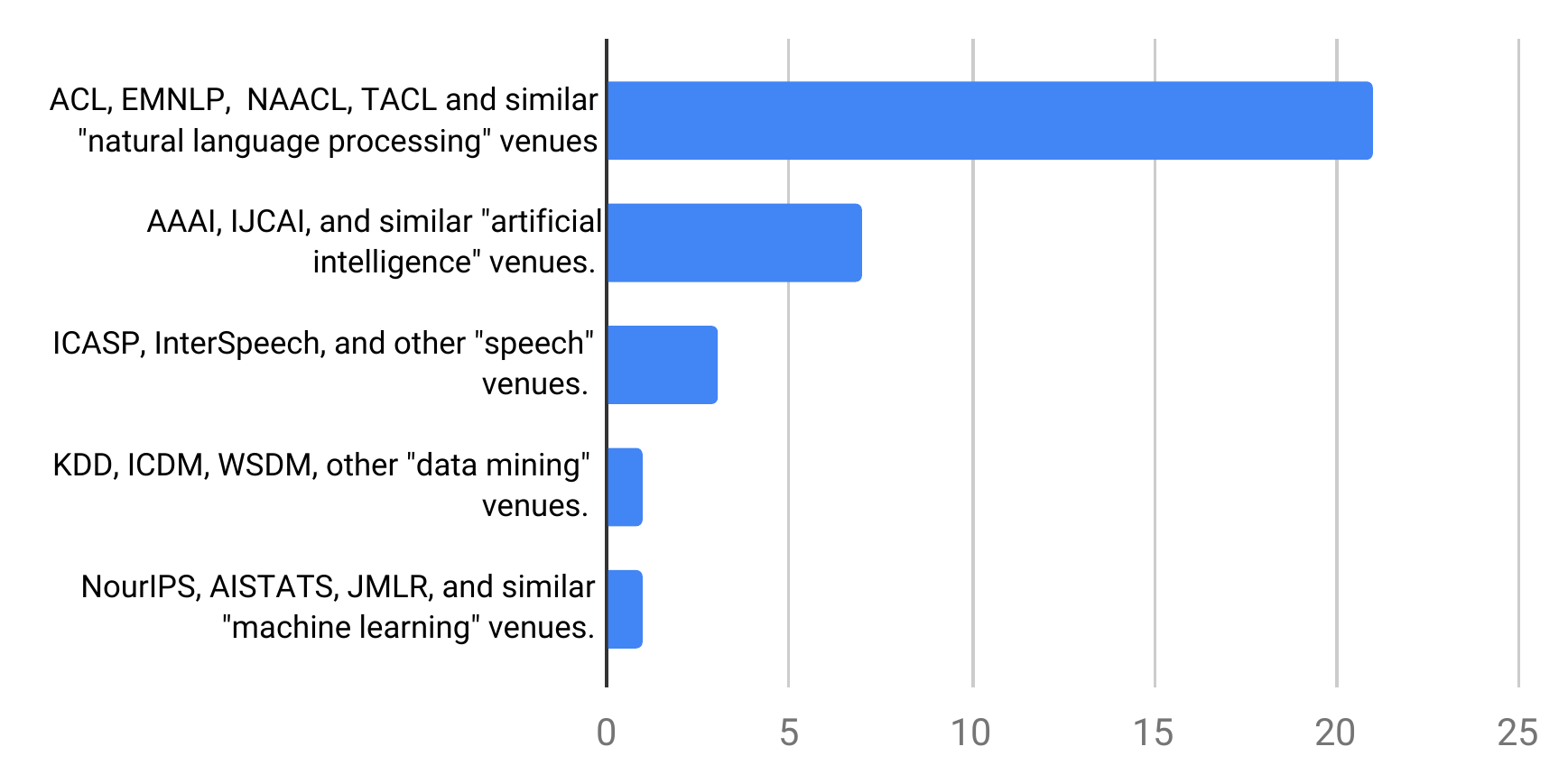}}
        \caption{Distribution of responses to ``What venues do you usually publish in?''}
        \label{fig:distribution_of_venues}
    \end{subfigure}
    
    \begin{subfigure}[t]{0.49\textwidth}
        \centering
        \raisebox{-\height}{\includegraphics[scale=0.26,trim=0cm 0.2cm 0.2cm 0cm, clip=false]{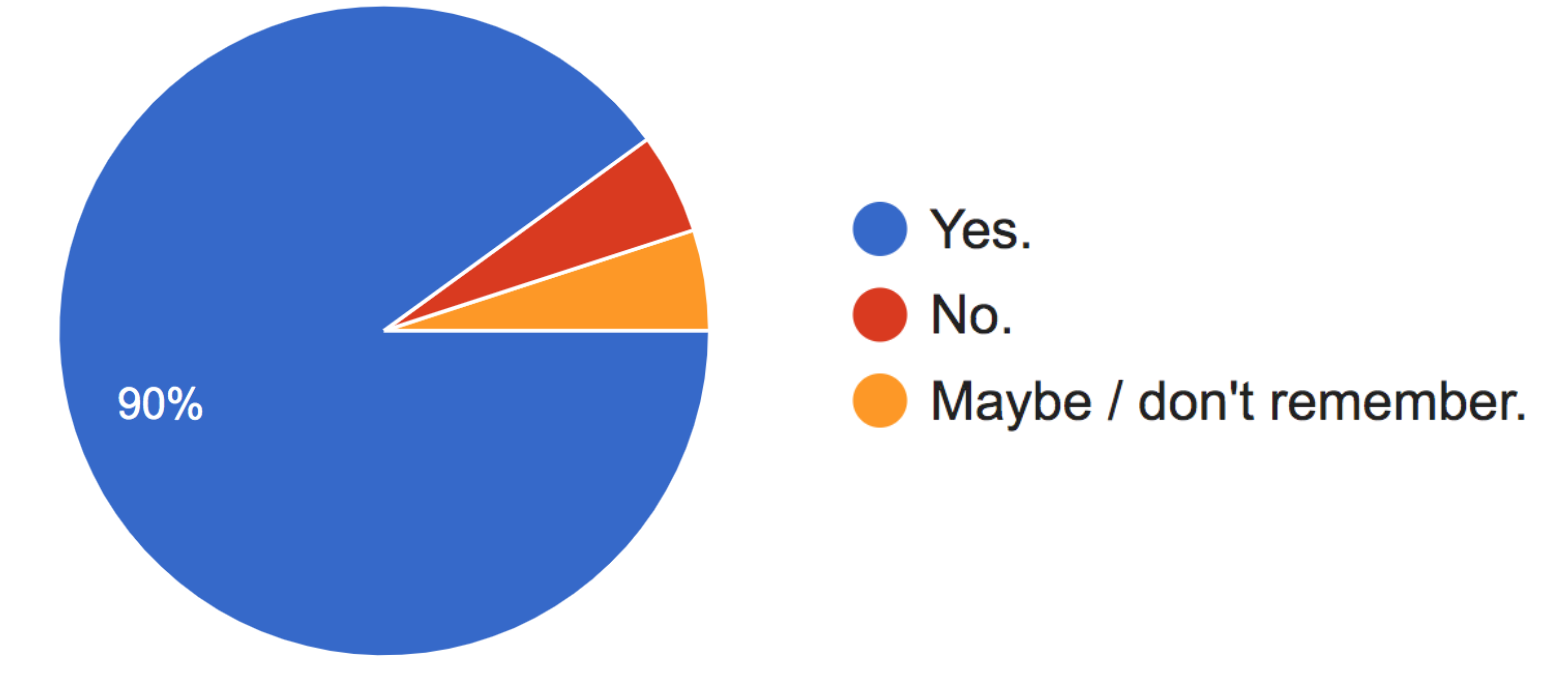}}
        \caption{Distribution of responses to ``I have learned about statistical hypothesis testing/assessment (via taking classes or reading it from other places).''}
    \label{fig:have:learned:significance:test}
    \end{subfigure}
    \hfill
    \begin{subfigure}[t]{0.49\textwidth}
        \centering
        \raisebox{-\height}{\includegraphics[scale=0.45,trim=0.23cm 0.45cm 0cm -0.5cm, clip=true]{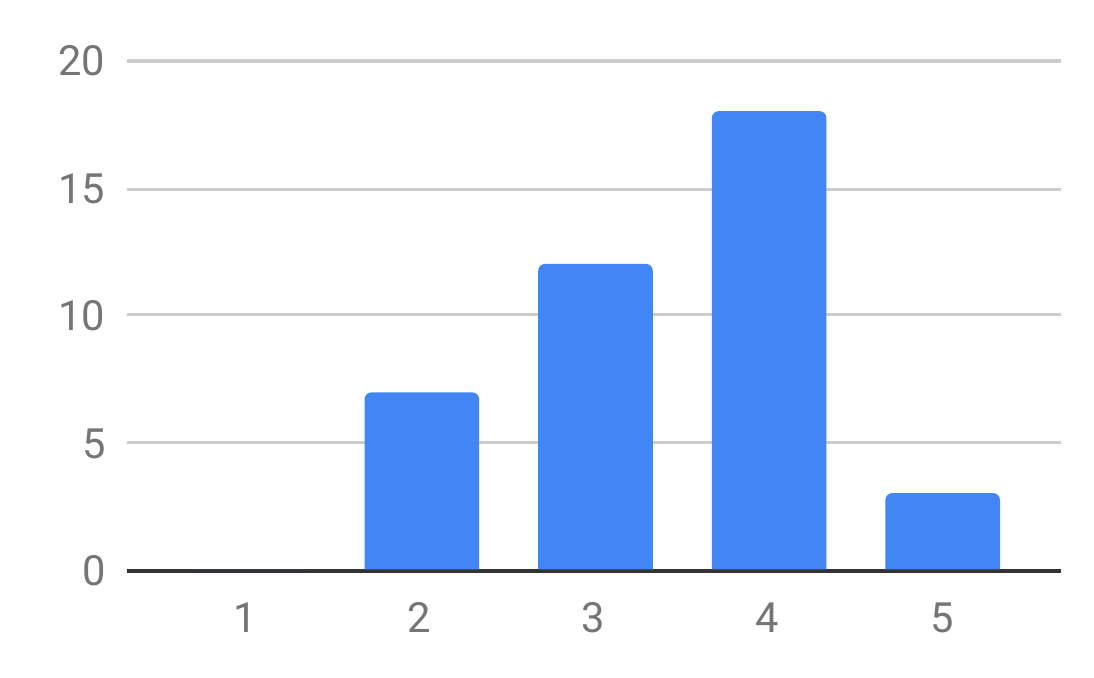}}
        \caption{Distribution of responses to ``I can understand almost all the "statistical" terms I encounter in papers.''}
        \label{fig:understand_statistical_terms}
    \end{subfigure}
    
    \begin{subfigure}[t]{0.49\textwidth}
        % \centering
        \raisebox{-\height}{\includegraphics[scale=0.28,trim=-3cm 0cm 0cm -1cm]{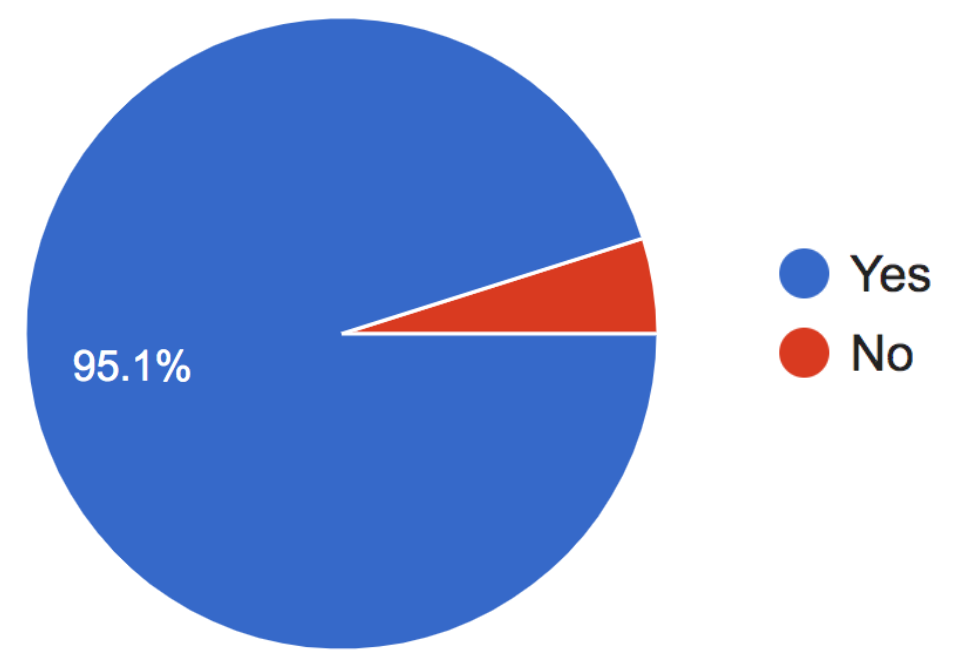}}
        \caption{Distribution of responses to ``Do you know the definition of Confidence Interval?''}
        \label{fig:know_confidence_intervals}
    \end{subfigure}
    \hfill
    \begin{subfigure}[t]{0.49\textwidth}
        \centering
        \raisebox{-\height}{\includegraphics[scale=0.45,trim=0.23cm 0.45cm 0cm -1cm, clip=true]{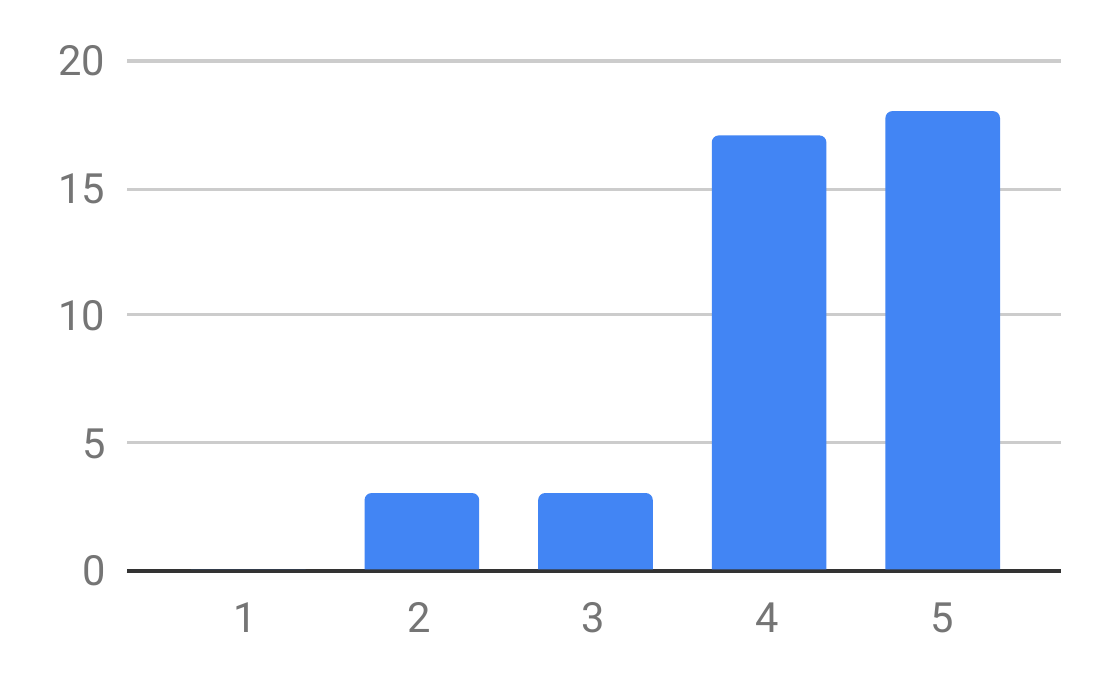}}
        \caption{Distribution of responses to ``I know $p$-values and I know how to interpret them.''}
        \label{fig:know_pvalues}
    \end{subfigure}
    \caption{Responses to several questions in our survey of NLP researchers}
    \label{figure:results:of:survey}
\end{figure*}

% \begin{figure*}[h]
% \centering
% \includegraphics[width=0.47\textwidth,trim=0cm 2.5cm 0cm 3cm, clip=false]{figures/BinomialQACompare_prior_theta.pdf}
% \includegraphics[width=0.47\textwidth,trim=0cm 2.5cm 0cm 3cm, clip=false]{figures/BinomialQACompare_posterior_theta.pdf}

% \caption{
% \small
% Left: Prior distributions of two systems (bottom row) and their difference (top row). 
% Right: 
% Posterior distributions of two systems (bottom row) and their difference (top row) after observing the performances on ARC-easy dataset.
% Here we assume at least one percent 
% % to be the threshold for the difference of accuracies 
% accuracy difference 
% to be considered practically different. 
% Hence, we indicate the interval $(-0.01, 0.01)$ to be the ROPE (introduced in \S\ref{subsec:posterior:interval}.) The interval is highlighted in red and the middle of the interval, indicating where the accuracies are exactly equal, is indicated with a green line in all figures.\footnote{The figure could be reproduced via the accompanying software.}
% }
% \label{fig:HDI2}
% \end{figure*}
% \input{calculations.tex}

\twocolumn

\clearpage

\section{Comparison, Misinterpretations, and Fallacies}
% \subsection{Fallacies Involving $p$-value}
% \section{Misinterpretations \& Malpractices of $p$-values}
% \paragraph{$p$-values malpractices.}
% \paragraph{$p$-values misinterpretations.}
\label{appendix:sec:comparisons}

This is an extension of the ideas discussed in Section~\ref{sec:comparisons}. 

\ignore{
When it comes to choosing an approach to assess significance of hypotheses, there are many issues that have to be taken into account. 
% Here we briefly go over a few of the well-known issues and compare the aforementioned techniques. 
This section compares different techniques in terms of various well-known issues. 

A summary of the techniques studied here is provided in Table~\ref{tab:method-comparison}. Notice that in the final column, $p$-value based techniques are dominant in the field. However, as we will delineate later, the alternatives could have advantages. 
}

\subsection{Susceptibility to Misinterpretation}
% \paragraph{Ambiguity in interpretation.}
% \paragraph{Ambiguity in interpretation.}
% \paragraph{Easiness to misinterpret.}
% \paragraph{Susceptibility to misinterpret.}
% \paragraph{Misinterpretation.}
\label{subsec:misinterpretation}

The complexity of interpreting significance tests, combined with insufficient reporting (as shown by~\newcite{dror2018hitchhiker}) 
% or the lack of familiarity with tests 
could result in ambiguous or misleading conclusions. 
Not only many papers are unclear about their tests, but their results could also be misinterpreted by readers. 

Among the techniques studied here, $p$-values, due to their complex definition, have received by far the biggest number of criticisms. 
While $p$-values are the most common approach, they are inherently complex which makes them easy to misinterpret. 
Here are a few common misinterpretations~\cite{demvsar2008appropriateness,goodman2008dirty}: 

\begin{itemize}[topsep=0.1pt,itemsep=0.1ex,partopsep=0.1ex,parsep=0.1ex,leftmargin=*]
    \item \textbf{Misconception \#1:} \emph{If $p < 0.05$, the null-hypothesis has only a 5\% chance of being true:} 
    % This is probably the biggest misconception about $p$-values. 
    To see that this is false, remember that $p$-value is defined with the assumption that null-hypothesis is correct (Eq.~\ref{equation:pvalue}.)
    \item \textbf{Misconception \#2:} \emph{If $p > 0.05$, there is no difference between the two systems:} Having large $p$-value only means that the null-hypothesis is consistent with the observations, but it does not tell anything about the likeliness of the null-hypothesis. 
    \item \textbf{Misconception \#3:} \emph{A statistically significant result ($p<0.05$) indicates a large/notable difference between two systems:} $p$-value only indicates strict superiority and provides no information about the 
    \emph{margin}
    % \emph{magnitude} (margin) 
    of the effect.
\end{itemize}
% Interested readers can refer to the existing critiques of $p$-values ~\cite{demvsar2008appropriateness,goodman2008dirty}. 

%The inherently complex notion of $p$-value makes it inaccessible.
% The inherent complexity of $p$-value's definition makes it inaccessible.
% CI can be seen as 
% an extension 
% a generalization 
% of the concept 
% of $p-$value. This means that working with CIs have even more complications added to the definition of $p-$value.
Interpretation of CI can also be challenging since it is an extension of $p$-value~\cite{hoekstra2014robust}. Tests that provide measures of uncertainty (like the ones in \S\ref{subsec:posterior:interval}) are more natural choices for reporting the results of experiments~\cite{kruschke2018bayesian}. Overall, our view on ``ease of interpretation'' of the approaches is summarized in Table~\ref{tab:method-comparison}. 

\subsection{Practical Significance}
 \label{subsec:practical_significance}
Statistical significance is different from practical significance~\cite{berger1987testing}. 
While we saw in Example~\ref{exm:p-value} that the difference between systems $S_1$ and $S_2$ is statistically significant (\textbf{\emph{H1}}), it does not provide any intuition on the \emph{magnitude} of their difference (the $x$ parameter in \textbf{\emph{H2-4}}). In other words, a small $p$-value does not necessarily mean that the effect is \emph{practically} important. Confidence intervals alleviate this issue by providing the range of parameters that are compatible with the data; however, they do not provide probability estimates~\cite{kruschke2018bayesian}. Bayesian analysis provides probability distributions directly over the target parameters of interest. This allows users to report uncertainty estimates for hypotheses that capture the margins of the effects. 

\subsection{Unintended Misleading Result by Iterative Testing}
% \paragraph{Unintended misleading result by multiple testing.}
While many tests are designed for a single-round experiment, in practice researchers perform multiple rounds of experiments until a predetermined condition is satisfied. This is particularly a problem in binary tests (such as $p$-value) when the condition is to achieve a desired result. For example, a researcher could continue experimenting until they achieve a statistically significant result (even if they don't necessarily have any intention of cheating the test)~\cite{kim2016three}. 

Since the outcomes of $p$-values and CIs only can ``reject'' or stay ``undecided'', these tests reinforce an unintentional bias towards the only possible decision. Consequently, it becomes easy to misuse this testing mechanism: for big enough data points it is possible to make statistically significant claims~\cite{amrhein2017earth}.

On the other hand, the approaches in \S\ref{subsec:posterior:interval} and \S\ref{subsec:bayes:factor} provide both outcomes of ``Accept'' and ``reject'', beside staying ``undecided.'' Therefore an honest researcher is more probable to accept that their data supports the opposite of what their conjecture was.
% A related known issue regarding $p-$values and CIs is that their definition depend on the \emph{stopping intention} of the researcher. It is common to perform the significance test with the assumption that there was a an agreed fixed sample size before performing the test. However in practice, this is rarely the case, as often times there are some other factors affecting the sample size, e.g., experiment is terminated when the researcher is satisfied the observation or gets tired to continue.

For an in-depth study of how each approach behaves in sequential tests, we refer to \citet[\S13.3]{kruschke2010bayesian}.  

Since $p$-values and CIs do not depend on prior they are not subject to this issue.

% \subsection{Sensitivity to the Choice of Prior.}
\label{subsec:priorSesitivity}
% The choice of prior could change the output of posteriors (both \S\ref{subsec:posterior:interval} and \S\ref{subsec:bayes:factor}). 
% It is a well-known issue that decisions based on Bayes Factor \S\ref{subsec:bayes:factor} are highly sensitive to the choice of prior, and less so the posterior estimates 
% in \S\ref{subsec:posterior:interval}~\cite{sinharay2002sensitivity,vanpaemel2010prior}. Since $p$-values and CIs do not depend on prior they are not subject to this issue.

\subsection{Choice of Prior for Bayes Factor}
\label{subsec:mitigateBFSens}
As discussed in \S\ref{subsec:bayes:factor}, \S\ref{subsec:priorSesitivity}, and \S\ref{sec:recommended_practices}, Bayes Factor is quite sensitive to the choice of prior. To address this, here are a few options to set the prior:
\begin{enumerate}
    \item Within the framework of model selection, if the priors are decided based on a clear meaning, as opposed to formulating a ``vague'' prior, then they can be justified to the audience.
    \item Often, there are a few choices of non-committal priors that seem equally representative of our beliefs. The best option is to perform and report one test for each of these choices to control the sensitivity to the prior.
    \item Another common approach to mitigate this concern is to use a small portion of the data to get an ``informed'' prior and do the analysis using this prior. This ensures that the new prior is meaningful and defensible.
    % \item Anotas answered er common approach to mitigate this concern is
    % use a small portion of the data to get an ``informed'' prior and do the analysis using this prior. This the Bayes Factor does not take extreme values 
    \item If none the above applies, it is recommended to use the approaches in \S\ref{subsec:posterior:interval} instead. Even though posterior density depends on prior, this measure is known to be robust for different choices of similar priors.
\end{enumerate}

\subsection{Flexible Choice of Hierarchical Model}
In this respect, the approaches in \S\ref{subsec:posterior:interval} and \S\ref{subsec:bayes:factor} are better suited to take the specifics of each setting into account.

\subsection{Bayes Factors vs.\ Posterior Intervals}
\label{subsec:bayes:factor:comparison}
We refer interested readers to \citet[pp.~165-166]{kruschke2018bayesian} for a list of Bayes factor caveats.

\end{document}